%% file: arxiv.tex
\documentclass{article} 
\usepackage{iclr2026_conference,times}

\input{math_commands.tex}

\usepackage{hyperref}
\usepackage{url}

\usepackage{dsfont}
\usepackage{graphicx} 

\usepackage[utf8]{inputenc} 
\usepackage[T1]{fontenc}    
\usepackage{hyperref}       
\usepackage{url}            
\usepackage{booktabs}       
\usepackage[linesnumbered,ruled,vlined]{algorithm2e} 
\usepackage{amsfonts}       
\usepackage{nicefrac}       
\usepackage{microtype}      

\usepackage{times}
\usepackage{latexsym}
\usepackage[table,xcdraw]{xcolor}
\usepackage{cuted}

\usepackage{makecell}
\usepackage{multirow}
\usepackage{hyperref}
\usepackage{url}
\usepackage{subcaption}
\usepackage{algorithmic}
\usepackage{amssymb}
\usepackage[most]{tcolorbox}
\usepackage{pifont}
\usepackage{color}
\usepackage{amsmath}
\usepackage{makecell}
\usepackage{newfloat}
\usepackage{listings}
\usepackage{afterpage}
\usepackage{colortbl} 
\usepackage{wrapfig}
\usepackage{soul}  
\usepackage{threeparttable}
\usepackage{array}    
\usepackage{ragged2e}  
\usepackage{longtable}
\usepackage{tabularx}

\usepackage{rotating}

\definecolor{wangfang}{RGB}{204,121,167}   
\definecolor{zhangyan}{RGB}{222,45,138}    
\definecolor{liuli}{RGB}{171,60,174}       

\definecolor{zhangjie}{RGB}{55,126,184}    
\definecolor{liwei}{RGB}{0,153,136}        
\definecolor{zhangtao}{RGB}{0,109,44}      

\newcommand{\WangFang}{\textcolor{wangfang}{\textbf{Wang Fang~}}}
\newcommand{\ZhangJie}{\textcolor{zhangjie}{\textbf{Zhang Jie~}}}
\newcommand{\ZhangYan}{\textcolor{zhangyan}{\textbf{Zhang Yan~}}}
\newcommand{\LiWei}{\textcolor{liwei}{\textbf{Li Wei~}}}
\newcommand{\LiuLi}{\textcolor{liuli}{\textbf{Liu Li~}}}
\newcommand{\ZhangTao}{\textcolor{zhangtao}{\textbf{Zhang Tao~}}}

\definecolor{c1}{HTML}{003371}
\definecolor{c2}{HTML}{057748}

\tcbuselibrary{breakable}
\NewDocumentCommand{\cdp}
{ mO{} }{\textcolor{blue}{\textsuperscript{\textit{Dong ping}}\textsf{\textbf{\small[#1]}}}}
\NewDocumentCommand{\yinuo}
{ mO{} }{\textcolor{blue}{\textsuperscript{\textit{Yi Nuo}}\textsf{\textbf{\small[#1]}}}}

\iclrfinalcopy
\title{EduVerse: A User-Defined Multi-Agent Simulation Space for Education Scenario}

\author{\textbf{Yiping Ma}$^{1*}$, \ 
\textbf{Shiyu Hu}$^{2*}$, \ 
\textbf{Buyuan Zhu}$^{2}$, \ 
\textbf{Yipei Wang}$^{3}$, \ 
\textbf{Yaxuan Kang}$^{4}$, \ \\
\textbf{Shiqing Liu}$^{5\dagger}$, \ 
\textbf{Kang Hao Cheong}$^{2,6\dagger}$\\
\textsuperscript{1}Lab of Artificial Intelligence for Education, East China Normal University,\\
\textsuperscript{2}School of Physical and Mathematical Sciences, Nanyang Technological University,\\
\textsuperscript{3}Institute of Automation, Southeast University\\
\textsuperscript{4}Institute of Automation, Chinese Academy of Sciences\\
\textsuperscript{5}Department of Education, East China Normal University\\
\textsuperscript{6}College of Computing and Data Science, Nanyang Technological University\\
\tt\small 52275901020@stu.ecnu.edu.cn \quad shiyu.hu@ntu.edu.sg \\
\tt\small buyuan001@e.ntu.edu.sg \quad 230248984@seu.edu.cn \quad yaxuan.kang@ia.ac.cn\\
\tt\small sqliu@dedu.ecnu.edu.cn  \quad kanghao.cheong@ntu.edu.sg\\
\footnotesize{$^{*}$ Equal Contribution \quad $^{\dagger}$ Corresponding Author}\\
}

\begin{document}

\maketitle

\begin{abstract}

Reproducing cognitive development, group interaction, and long-term evolution in virtual classrooms remains a core challenge for educational AI, as real classrooms integrate open-ended cognition, dynamic social interaction, affective factors, and multi-session development rarely captured together. Existing approaches mostly focus on short-term or single-agent settings, limiting systematic study of classroom complexity and cross-task reuse.
We present \textbf{EduVerse}, the first \emph{user-defined} multi-agent simulation space that supports environment, agent, and session customization. A distinctive human-in-the-loop interface further allows real users to join the space. Built on a layered \textbf{CIE} (\textbf{C}ognition–\textbf{I}nteraction–\textbf{E}volution) architecture, EduVerse ensures individual consistency, authentic interaction, and longitudinal adaptation in cognition, emotion, and behavior—reproducing realistic classroom dynamics with seamless human–agent integration.
We validate EduVerse in middle-school Chinese classes across three text genres, environments, and multiple sessions. Results show: \textbf{(i) Instructional alignment}: simulated IRF rates ($0.28$--$0.64$) closely match real classrooms ($0.37$--$0.49$), indicating pedagogical realism; \textbf{(ii) Group interaction and role differentiation}: network density ($0.27$–$0.40$) with about one-third of peer links realized, while human–agent tasks indicate a balance between individual variability and instructional stability; \textbf{(iii) Cross-session evolution}: the positive transition rate $R^{+}$ increase by 11.7\% on average, capturing longitudinal shifts in behavior, emotion, and cognition and revealing structured learning trajectories.
Overall, EduVerse balances realism, reproducibility, and interpretability, providing a scalable platform for educational AI. The system will be open-sourced to foster cross-disciplinary research.
\end{abstract}

\section{Introduction}
\label{sec:intro}

\begin{figure}[htbp]
    \centering
    \includegraphics[width=\textwidth]{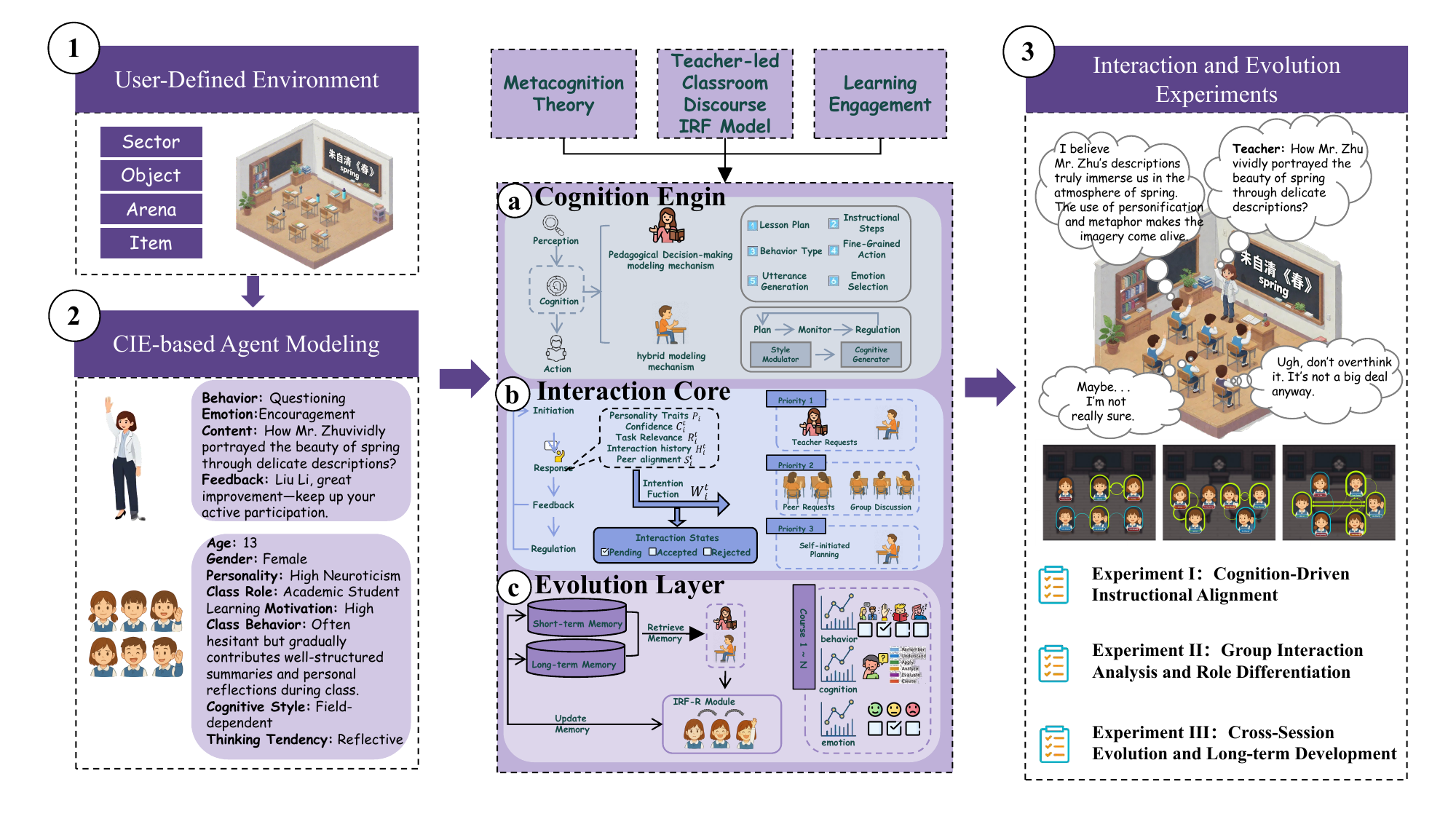}
    \caption{\textbf{Overview of EduVerse.} The framework consists of three main components: 
    (i) user-defined environment configuration (including classroom layouts, seating arrangements, and interaction networks); 
    (ii) CIE-based agent modeling (a three-layer Cognition–Interaction–Evolution architecture for teacher and student agents); 
    (iii) interaction and evolution experiments (covering instructional alignment, group interaction, and cross-session development). 
    Together, these components form a scalable, interpretable, and transferable multi-agent simulation platform for educational AI.
    }
    \label{fig:eduverse_framework}
\end{figure}

A central challenge in human-centered AI is to simultaneously reproduce cognitive development, group interaction, and long-term evolution within virtual environments \citep{wang2024comprehensivesurveycontinuallearning, parisi_continual_2019, zheng2024lifelonglearninglargelanguage, chen_lifelong_nodate, zheng2025lifelonglearninglargelanguage}. While large language models (LLMs) excel at language understanding and immediate task completion, most research remains confined to static tasks or short-term interactions, falling short of capturing evolving cognition, stable behavioral styles, and socially dynamic processes \citep{maharana2024evaluatinglongtermconversationalmemory, wang2025recursivelysummarizingenableslongterm, tan2025prospectretrospectreflectivememory, li2025helloagainllmpoweredpersonalized}. Similarly, multi-agent systems have primarily targeted structured games or fixed collaboration, lacking frameworks that support developmental agents whose cognition, personality, and social relations evolve naturally over time \citep{Wang_2024, ashery2024dynamicssocialconventionsllm}.

Educational settings, particularly classrooms, provide an ideal testbed for addressing these challenges. Classroom dynamics inherently integrate knowledge construction, individual differences, social interaction, and real-time instructional feedback \citep{doi:10.3102/003465430298487, poropat_meta-analysis_2009, Johnson01071998}. For instance, Chinese language classes, with open-ended tasks, rich emotional expression, and dense role-based interactions, offer fertile ground for modeling cognitive development and group dynamics. However, existing intelligent tutoring systems, personalized assistants, and generative dialogue agents \citep{Anderson01041995, nye_autotutor_2014, lin_artificial_2023} largely treat students as static performers rather than adaptive learners. They lack mechanisms for persistent individual modeling, role-differentiated social interaction, and longitudinal instructional adaptation, leaving developmental trajectories underrepresented.

To overcome these limitations, we introduce \textbf{EduVerse}, the first \emph{user-defined} multi-agent simulation space that supports environment customization through flexible physical layouts and seating arrangements, agent customization via a human-in-the-loop interface integrated with a layered \textbf{CIE} (\textbf{C}ognition–\textbf{I}nteraction–\textbf{E}volution) architecture, and session customization for modeling multi-lesson trajectories. In CIE, the cognition layer ensures individual consistency and instructional alignment, the interaction layer models priority-based authentic exchanges, and the evolution layer captures longitudinal changes in cognition, emotion and behavior. Together, these capabilities enable EduVerse to reproduce realistic classroom dynamics while supporting seamless human–agent interaction.

We instantiate EduVerse in middle school Chinese language classes, characterized by open-ended tasks, rich emotional expression, and complex interaction structures. Through three experiments aligned with the CIE dimensions, we examine: \textbf{(i) Instructional alignment}: simulated IRF rates ($0.28$--$0.64$) closely match real classrooms ($0.37$--$0.49$), indicating pedagogical realism; \textbf{(ii) Group interaction and role differentiation}: network density ($0.27$–$0.40$) with about one-third of peer links realized, while human–agent tasks indicate a balance between individual variability and instructional stability; \textbf{(iii) Cross-session evolution}: the positive transition rate $R^{+}$ increase by 11.7\% on average, capturing longitudinal shifts in behavior, emotion, and cognition and revealing structured learning trajectories. These studies demonstrate EduVerse’s ability to reproduce both individual and group dynamics while uncovering multi-dimensional learning trajectories.

\textbf{Contributions.}  
(1) We propose EduVerse, the first user-defined multi-agent simulation space for education, enabling reusable and customizable experimentation across tasks and disciplines.  
(2) We design the CIE architecture, which systematically models the cognitive, interactive, and evolutionary dynamics of developmental agents.  
(3) We conduct instantiated experiments and analyses, demonstrating EduVerse’s potential in authentic educational contexts.  
As an extensible open framework, EduVerse redefines virtual classroom modeling and provides a systematic and cross-disciplinary pathway for educational AI; it will be open-sourced to foster transparency and collaboration.

\section{Related Work}
\label{sec:related-work}

\textbf{EduVerse} aims to build a user-defined multi-agent simulation space for education that systematically supports cognitive development, group interaction, and long-term evolution in virtual classrooms. Despite progress in educational agents, multi-agent simulation, and LLM-based generation, a unified platform integrating these dimensions remains absent. We therefore review three threads closely aligned with EduVerse’s dimensions (see App.~\ref{sec:more-relate-work} for details).

\textbf{Educational agents and virtual classrooms.}
Early systems such as \textit{Cognitive Tutor} \citep{Anderson01041995} and \textit{SimStudent} \citep{article} focused on skill acquisition and personalization, typically via rule- or model-based mechanisms \citep{christensen2011simschool, foley2005making, carrington2011enhancing, dotger2010medicine}. Teacher-training simulations used scripted virtual students as scaffolds but lacked adaptivity to feedback, peer influence, or classroom context \citep{kervin2006classsim, dieker2015tle, delamarre2021interactive, shernoff2018early, kelleci2021using}. Recent generative extensions enable task-level learning \citep{zhang_simulating_2024, lee_generative_nodate, yue_mathvc_2025, mollick_ai_2024, markel_gpteach_2023, wang_generative_2025, fahid_online_2024}, but often omit emotional modeling, stylistic progression, and multi-agent coupling, limiting their suitability for open, dynamic classrooms.

\textbf{Multi-agent social simulations.}
Works such as \textit{Generative Agents} show that LLMs enhanced with memory, planning, and reflection can generate human-like social behaviors in sandbox settings \citep{park2023generative, li2023camelcommunicativeagentsmind, chen2023agentversefacilitatingmultiagentcollaboration, jinxin_cgmi_2023}. However, these focus on adult roles and informal contexts, overlooking classroom-specific structures such as IRF discourse, teacher–student roles, and goal alignment. They also lack mechanisms for knowledge progression tracking and temporal adaptivity.

\textbf{Personalized modeling and long-term coherence.}
Persona conditioning and style control are widely used to maintain role consistency \citep{shao_character-llm_2023, jiang2024personallminvestigatingabilitylarge, wang2024rolellmbenchmarkingelicitingenhancing}, with design patterns surveyed by \citet{tseng-etal-2024-two}. Yet long-term interactions often suffer from persona drift, leading to memory-based prompting \citep{zhong2023memorybankenhancinglargelanguage}, style constraints \citep{roy2023conversationstyletransferusing}, and metacognitive or reflective mechanisms \citep{madaan2023selfrefineiterativerefinementselffeedback, li_exploring_2025, didolkar2024metacognitivecapabilitiesllmsexploration}. Research on continual and lifelong learning also contributes to longitudinal coherence \citep{wang2024comprehensivesurveycontinuallearning, parisi_continual_2019, zheng2024lifelonglearninglargelanguage, chen_lifelong_nodate, zheng2025lifelonglearninglargelanguage, maharana2024evaluatinglongtermconversationalmemory, wang2025recursivelysummarizingenableslongterm, tan2025prospectretrospectreflectivememory, li2025helloagainllmpoweredpersonalized}. However, these methods are mostly evaluated in single-agent or non-classroom contexts, rarely integrating group-level structures or pedagogically grounded evolution. \textit{EduAgent} \citep{xu_eduagent_2024} models individual cognitive and metacognitive processes but lacks multi-agent coordination and group dynamics.

Overall, Prior work offers key ingredients—feedback, generative behavior, and role consistency—but remains fragmented across time spans (short-term vs. longitudinal), modeling dimensions (individual vs. group), and educational contexts (informal vs. classroom-structured). \textbf{EduVerse} bridges these gaps by combining user-defined environments, agent, sessions and by adopting the \textbf{CIE} architecture to unify individual coherence, group dynamics, and cross-session evolution in a scalable and interpretable classroom simulation platform.

\section{EduVerse Framework}
\label{sec:eduverse}

\textbf{EduVerse} is a user-defined multi-agent simulation framework for educational settings, designed to model the long-term cognitive, behavioral, and social interactions of \textit{developing learners}. Its design integrates three complementary components: (i) The \textbf{user-defined environment} for configuring classroom layouts, seating, and interaction networks to generate diverse instructional scenarios within a unified physical–social space (see Sec.~\ref{subsec:environment}); (ii) the \textbf{CIE-based agent modeling} equips student agents with a unified perception–cognition–action (PCA) architecture, augmented by personalized embeddings and style modulation for both cognitive coherence and expressive diversity, while a human-in-the-loop interface enables user-defined customization of agents (see Sec.~\ref{subsec:agent}), and (iii) the \textbf{interaction and evolution experiments} combine teacher-led guidance with student-initiated behaviors to examine instructional alignment, group interaction, and cross-session evolution (see Sec.~\ref{subsec:experiment}). 
As illustrated in Fig.~\ref{fig:eduverse_framework}, these components together form a scalable, interpretable, and transferable foundation for subsequent analyses.

\subsection{User-defined Environment}
\label{subsec:environment}

The environment module constructs an interaction space that integrates \emph{physical constraints} with \emph{social semantics}. We adopt a hierarchical spatial structure
$\mathcal{Z}=\{\mathcal{Z}_S,\mathcal{Z}_A,\mathcal{Z}_O,\mathcal{Z}_I\}$:
$\mathcal{Z}_S$ denotes functional \emph{sectors} (e.g., teacher, student, activity zones),
$\mathcal{Z}_A$ denotes localized \emph{arenas} (e.g., a discussion circle or a podium),
$\mathcal{Z}_O$ denotes interactive \emph{objects} (e.g., blackboards, podiums, desks), and
$\mathcal{Z}_I$ represents fine-grained \emph{items} (e.g., textbooks, pens, chalk).
This layered organization provides a unified mapping between physical distribution and pedagogical semantics, consistent with App.~\ref{subapp:environment}.

Peer interaction is captured by a seat-adjacency graph
$A^{seat}\in\{0,1\}^{N\times N}$ defined as
\begin{equation}
A^{seat}_{ij}=
\begin{cases}
1,& \text{if students $i$ and $j$ satisfy the adjacency rules},\\
0,& \text{otherwise},
\end{cases}
\end{equation}
where the rules may combine distance $d(i,j)$, group membership $g(i)$, and layout-specific constraints.
Researchers can instantiate different interaction topologies by editing configuration files, avoiding hard-coded seat links.

\begin{figure}[ht!]
\centering
\includegraphics[width=\textwidth]{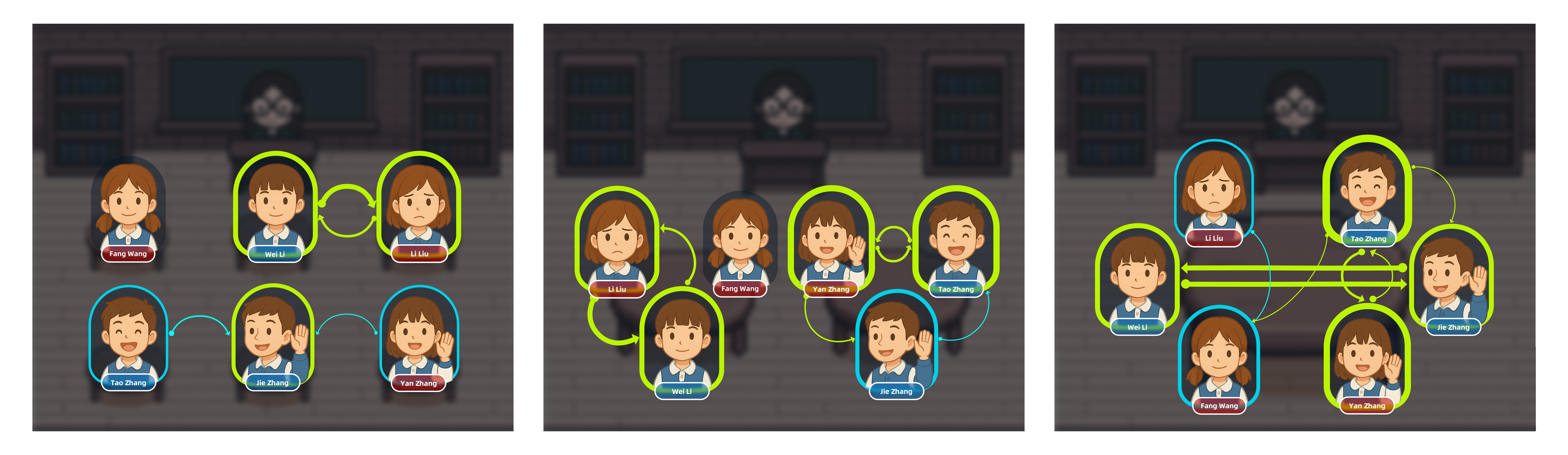}
\caption{\textbf{Visualization of student group interactions across classroom environments.}
    The three panels correspond to: Lecture (left, traditional teacher-centered layout), 
    Collab\_Two\_Tables (middle, two-group collaborative setting), and 
    Round\_Table (right, open discussion layout).
    This figure highlights EduVerse’s capability to vary classroom environments, while showing that the resulting interaction topologies naturally support subsequent analyses of group dynamics and evolution experiments (see Sec.~\ref{subsec:exp-i}).}
\label{fig:student-interaction}
\end{figure}

As shown in Fig.~\ref{fig:student-interaction}, under this unified definition, EduVerse implements three canonical classroom layouts:
\textbf{Lecture} constrains peer links by distance and group, reflecting teacher-centered, largely unidirectional communication;
\textbf{Round\_Table} augments distance-based adjacency with face-to-face \emph{opposite-seat} edges $j=\operatorname{opp}(i)$ to encourage open peer dialogue;
\textbf{Collab\_Two\_Tables} forms fully connected within-group subgraphs with no cross-group links, emphasizing intra-group collaboration and bounded social structure.

These layouts serve as illustrative cases rather than limitations.
Users can freely customize the hierarchy $\mathcal{Z}$ and the adjacency-generation rules via configuration files to simulate classrooms of varying scales, tasks, and pedagogical styles.
Leveraging the \texttt{seat\_graph} mechanism, EduVerse tightly couples physical space with social semantics, providing a \emph{realistic, controllable, and extensible} environment foundation for subsequent agent decision-making and group-level experiments.

\subsection{CIE-based Agent Modeling}
\label{subsec:agent}

Agent modeling in \textbf{CIE} is grounded in a unified PCA loop and organized along three dimensions: \textbf{cognition-driven decision-making}, \textbf{social situatedness and group coordination}, and \textbf{cross-lesson temporal dynamics}. Together, these mechanisms connect the levels of individual, group, and time, forming the basis for modeling developing learners in classroom simulations (see App.~\ref{subapp:cie-c}--\ref{subapp:cie-m}).

\subsubsection{Cognition-Driven Agent Decision Mechanism}
\label{subsubsec:cie-c}

All agents $\mathcal{A}_i$ follow the PCA architecture, formalized as:
\begin{equation}
\mathcal{A}_i^t:(\mathcal{O}_i^t,\mathbf{e}_i)\rightarrow a_i^t,
\label{eq:decision}
\end{equation}
where $\mathcal{O}_i^t$ is the local observation extracted from the global state $\mathcal{S}^t$ via a perception function $\mathcal{P}_i$, and $\mathbf{e}_i=[\mathbf{p}_i;\mathbf{c}_i;\mathbf{m}_i]$ encodes personality traits, cognitive style, and motivation. The action is generated by a language model $a_i^t = f_{\text{LLM}}(\mathcal{O}_i^t,\mathbf{e}_i)$.

We adopt a lightweight interaction gate to decide whether an agent takes a turn at time $t$:
\begin{equation}
\mathcal{G}_i^t=
\begin{cases}
1,& \text{if the scheduler assigns a teacher- or peer-directed turn to agent $i$ at time $t$},\\
0,& \text{otherwise (self-initiated behaviors may be scheduled separately).}
\end{cases}
\label{eq:gating}
\end{equation}
The cognitive loop consists of \emph{planning, monitoring, and regulation}; the gate $\mathcal{G}_i^t$ determines whether the loop is executed at time $t$.

\textbf{Content--style separation.}
To balance long-term consistency with expressive diversity, we decouple semantic planning from stylistic expression. Inspired by style transfer~\citep{gatys2016image,deng2022stytr2}, we adopt a two-component design:
(1) a \emph{style modulator} fine-tuned on educational data with InternVL~\citep{chen2024internvl} that produces style-aware prompts conditioned on traits and task phase (e.g., hesitancy, verbosity, affective tone), rather than direct responses; and
(2) a \emph{cognitive generator} that integrates the style prompt with $\mathcal{O}_i^t$, $\mathbf{e}_i$, and dialogue history to form a composite prompt for GPT-4~\citep{achiam2023gpt}, which focuses on semantic planning and content generation.
This design improves \textbf{ cognitive coherence} and \textbf{expressive diversity}, while offering an interpretable path for personality-conditioned behaviors.

\textbf{Role-specific parameterization.}
While student agents $\mathcal{A}_S^i$ and the teacher agent $\mathcal{A}_T$ share the PCA backbone, they differ in input channels, gating, and prompting. As shown in Tab.~\ref{tab:agent-diff}, students are modulated by personality and willingness embeddings, whereas the teacher relies on scripted lesson plans and classroom metrics. This unified yet differentiated design supports dynamics from teacher-led instruction to student-initiated contributions.

\subsubsection{Social Situatedness: Group Interaction and Behavioral Coordination}
\label{subsubsec:cie-i}

To better capture classroom discourse, CIE extends the classic IRF (Initiation–Response–Feedback) cycle to 
$I_T^t \rightarrow R_S^t \rightarrow F_T^t \rightarrow \text{Regulate}_S^t$.
At time $t$, the teacher agent $\mathcal{A}_T$ initiates a task $T^t$; student agents $\mathcal{A}_S^i$ respond, receive feedback, and then regulate subsequent actions, forming a micro-cycle aligned with instructional goals. Action selection follows Eq.~\ref{eq:decision} and is modulated by the interaction gate $\mathcal{G}_i^t$ (see Eq.~\ref{eq:gating}) to determine participation.

\textbf{Intention function.} We conceptualize willingness and responsiveness as 
$\omega_i^t = \alpha_1 P_i + \alpha_2 C_i^t + \alpha_3 R_i^t + \alpha_4 H_i^t + \alpha_5 S_{ij}^t$,
where $P_i$ denotes personality, $C_i^t$ confidence at time $t$, $R_i^t$ task relevance, $H_i^t$ interaction history, and $S_{ij}^t$ alignment with peer $j$; $\alpha_k$ are weighting coefficients. 

In practice, this abstraction is embedded into prompt design and scheduling to guide gating and response generation. Beyond teacher-led turns, students may self-initiate behaviors (e.g., questioning, head-up), enabling multi-party interaction and group discussion. The teacher monitors these dynamics and provides targeted or global feedback, closing the loop between pedagogical intent, social context, and adaptive regulation for analyzing group coordination and evolution.

\subsubsection{Temporal Dynamics: Cross-Lesson Adaptation and Evolution}
\label{subsubsec:cie-e}

CIE supports long-term, cross-lesson simulation through four mechanisms: knowledge progression, behavioral style regulation, instructional pacing control, and memory interaction flow.

\textbf{Knowledge progression.}
Each student agent $\mathcal{A}_S^i$ initializes a knowledge state $\mathbf{s}_i^0$ derived from $\mathbf{e}_i$, which evolves as $\mathbf{s}_i^{t+1}=\mathcal{R}_i(\mathbf{s}_i^t,a_i^t,F_i^t),\ \{\mathbf{s}_i^t\}_{t=1}^{T}$,
where $\mathcal{R}_i$ adjusts state components (e.g., confidence, engagement) using annotated behavioral signals (e.g., Bloom level, response type) and structured feedback (positive, neutral, negative).

\textbf{Behavioral style regulation.}
At each step, actions, reflections, affect, and cognitive states are logged in a structured growth log (JSON), enabling the tracking of style stability, engagement shifts, and recovery. These logs also feed subsequent scheduling and adaptation.

\textbf{Instructional pacing.}
The teacher agent organizes instruction into \textit{phases}, each comprising multiple \textit{steps}. After each phase, pacing is updated as $\text{Phase}_{k+1}\leftarrow \text{Transition}(\text{Phase}_k,\{\mathbf{s}_i^t\},\text{completion rate})$,
with policy $\pi_T$ dynamically adjusting step granularity, tone, and targeting. In practice, cycles are initialized with 30 steps and then adapt to interaction dynamics, forming a closed loop for adaptive teaching control.

\textbf{Memory interaction flow.}
To sustain cross-session continuity, CIE coordinates short- and long-term memory: long-term summaries are loaded at lesson start, short-term states are updated in real time, and aggregated records are written back after each session. This flow enables feedback-driven self-regulation and coherent developmental trajectories across lessons.

Taken together, the PCA backbone of CIE, coupled with mechanisms for cognition-driven decision-making, group interaction, and temporal evolution, yields behaviors that are \emph{stable yet diverse} within a lesson and \emph{coherent} across lessons, providing a robust foundation for subsequent experiments and pedagogical analyses.

\subsection{Interaction and Evolution Experiments}
\label{subsec:experiment}

This subsection formalizes EduVerse’s experimental paradigm. By configuring environments, agents, and cross-session evolution, researchers can design studies that probe cognitive alignment, group interaction, and long-term development. Concretely, they may specify the physical–social environment (e.g., seating layouts, interaction networks), configure agents with distinct personalities and cognitive styles, and define task flows and instructional scripts. On this basis, three core categories of experiments can be defined:
(i) Cognition-driven instructional alignment experiments;
(ii) Group interaction analysis and role differentiation experiments;
(iii) Cross-session evolution and long-term development experiments.
All experiments are logged and evaluated with unified metrics (e.g., IRF discourse structure, network density/centrality, positive transition rate $R^{+}$), ensuring interpretability and reproducibility.

In addition, EduVerse provides a \emph{human-in-the-loop} interface that admits real students or teachers alongside virtual agents, enabling simulation, causal testing, and validation within a single, low-cost, controllable, and interpretable framework. In Sec.~\ref{sec:experiment}, we demonstrate these capabilities in a \textbf{junior secondary Chinese classroom}, highlighting EduVerse’s applicability and research value.

\section{Experimental Design and Evaluation}
\label{sec:experiment}

\begin{wraptable}{r}{0.6\textwidth}
  \centering
  \vspace{-10pt}
  \caption{Average IRF (Initiation--Response--Feedback) distributions in simulated environments vs.~real classrooms across three text genres.}
  \label{tab:avg-IRF}
  \vspace{-5pt}
  \setlength{\tabcolsep}{4pt}
  \small
  \begin{tabular}{p{2cm}lcccc}
    \toprule
    \textbf{Genre} & \textbf{Setting} & \textbf{I} & \textbf{R} & \textbf{F} & \textbf{IRF\_rate} \\
    \midrule
    \multirow{2}{*}{\textit{Lyrical Prose}}       
      & Simulation & 0.454 & 0.166 & 0.293 & 0.336 \\
      & Real Class & 0.513 & --    & 0.703 & 0.486 \\
    \midrule
    \multirow{2}{2.5cm}{\textit{Argumentative Essay}} 
      & Simulation & 0.482 & 0.207 & 0.335 & 0.554 \\
      & Real Class & 0.417 & --    & 0.583 & 0.417 \\
    \midrule
    \multirow{2}{*}{\textit{Foreign Fiction}}     
      & Simulation & 0.310 & 0.230 & 0.407 & 0.379 \\
      & Real Class & 0.367 & --    & 0.515 & 0.367 \\
    \bottomrule
  \end{tabular}
  \vspace{-10pt}
\end{wraptable}

To demonstrate EduVerse’s customizability, we design three experiments on \textbf{cognitive alignment}, \textbf{group interaction}, and \textbf{long-term development}. Each adopts a \emph{teacher-led} mode with one teacher and multiple students, instantiated through three genres: lyrical prose (\textit{Spring}), foreign fiction (\textit{The Emperor’s New Clothes}), and argumentative essays (\textit{Dedication and Joy}). We select junior secondary Chinese instruction as the context, since its language-rich and discourse-driven nature provides a realistic and challenging setting for multi-turn IRF interactions and structured lesson phases. Implementation details appear in App.~\ref{sec:detailed-exp}. The three experiments align with EduVerse’s core capacities:
(1) \textbf{Experiment I} (Sec.~\ref{subsec:exp-c}): evaluates classroom authenticity and personality alignment under customized environment settings;  
(2) \textbf{Experiment II} (Sec.~\ref{subsec:exp-i}): investigates group interaction and individual influence, and validates human–agent interaction through the open agent interface;  
(3) \textbf{Experiment III} (Sec.~\ref{subsec:exp-e}): tracks students’ behavioral, emotional, and cognitive trajectories across four consecutive lessons to illustrate cross-session customization.
All experiments adopt a unified hybrid setup that combines trait-conditioned control with LLM-based generation, ensuring both behavioral consistency and interpretability. Detailed definitions of evaluation metrics are provided in App.~\ref{sub:evaluation-matrix}. We instantiate 6 virtual students (3 male and 3 female) with names drawn from common Chinese naming conventions without any prescriptive implications (See detailes in App.~\ref{subsec:agent})

\subsection{Experiment I: Environment Customization for cognition-driven instructional alignment}
\label{subsec:exp-c}

\begin{figure}[t]
    \centering
    \begin{subfigure}[b]{0.48\textwidth}
        \centering
        \includegraphics[width=\textwidth]{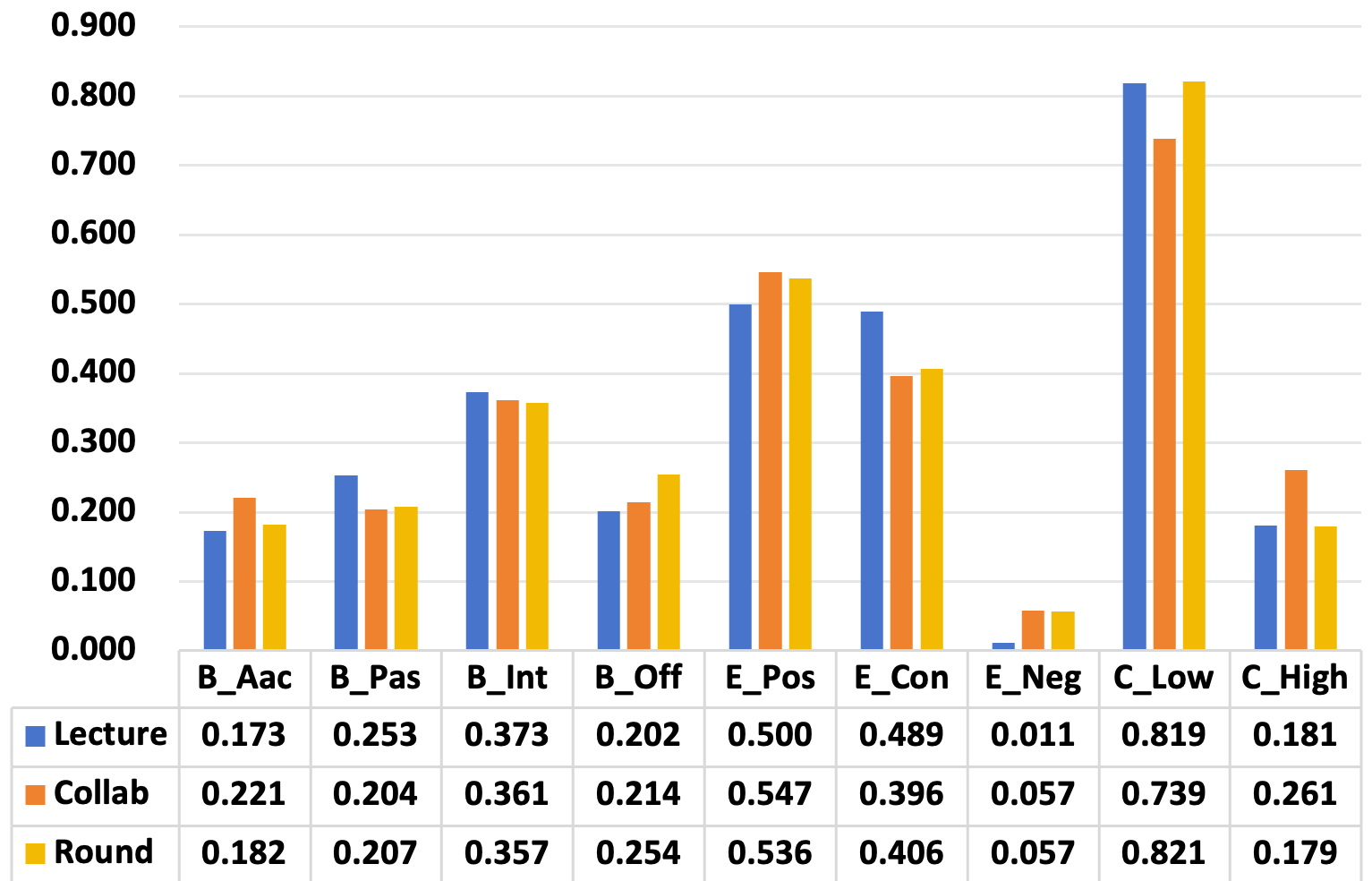}
        \caption{\textbf{Distribution of students’ BEC across environments.} 
        Different layouts yield distinct patterns: collaborative fosters positivity and higher-order cognition, round-table increases disengagement, while lecture remains stable with passive, lower-order patterns.}
        \label{fig:students-behavior-distribution}
    \end{subfigure}
    \hfill
    \begin{subfigure}[b]{0.48\textwidth}
        \centering
        \includegraphics[width=\textwidth]{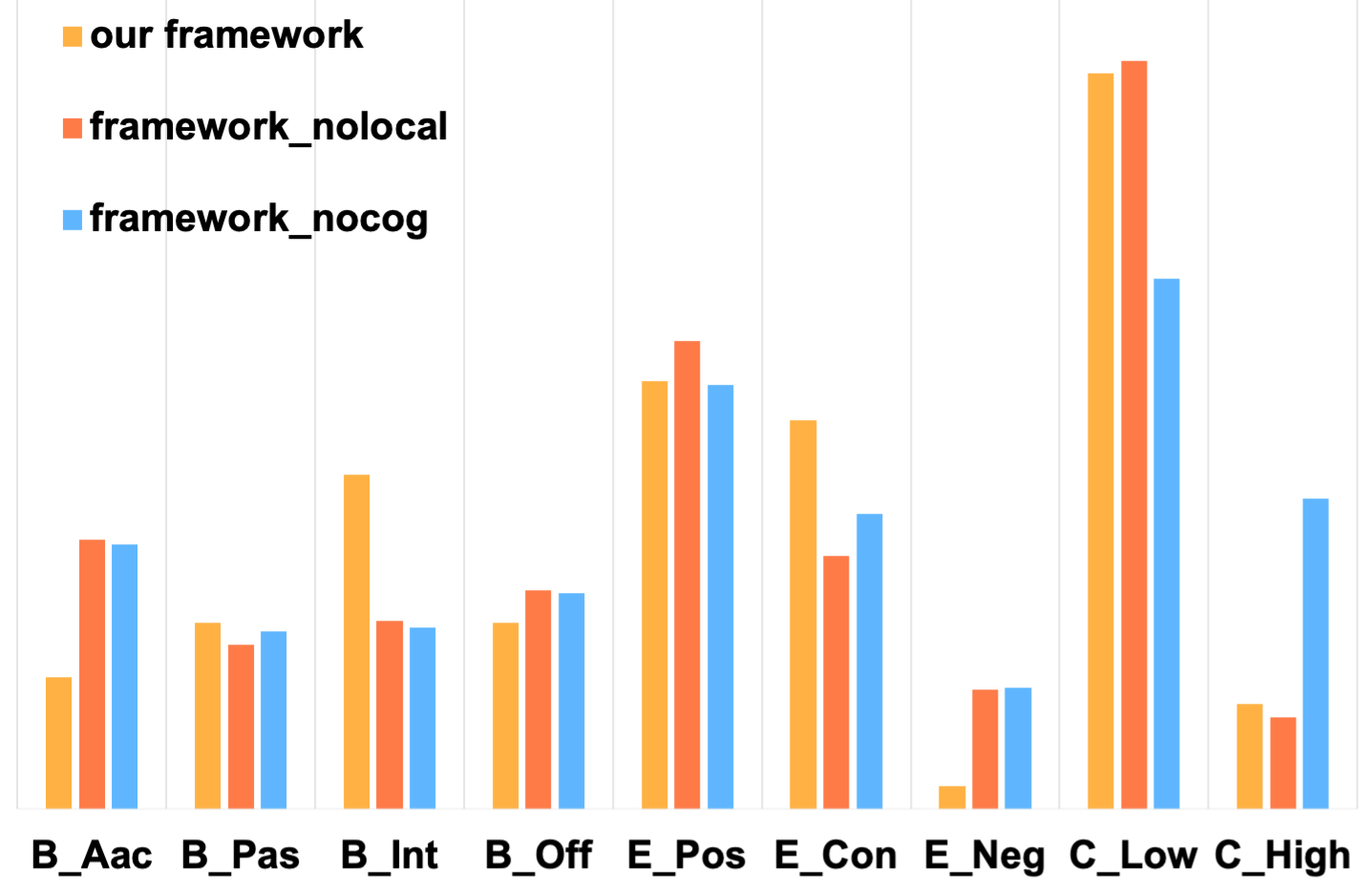}
        \caption{\textbf{Ablation study.} 
        Removing the stylization module amplifies negative emotions, while removing the CIE module exaggerates higher-order cognition and overly active behaviors.}
        \label{fig:Ablation_study}
    \end{subfigure}
    \caption{\textbf{Overall analysis of student behavior and system design.} 
    (a) Distribution of students’ behavioral–emotional–cognitive (BEC) patterns across different classroom environments. 
    (b) Ablation study illustrating the distinct roles of stylization and CIE modules.}
    \label{fig:student-behavior-ablation}
\end{figure}

This experiment examines whether customized environments can reproduce realistic classroom dynamics while preserving personality-driven behaviors. We instantiate three user-defined environments (Sec.~\ref{subsec:environment}) under a teacher-led mode and conduct ablations to evaluate key module contributions.

\textbf{IRF discourse patterns.}
As shown in Tab.~\ref{tab:avg-IRF}, the average IRF rates of simulated classrooms (0.336–0.554) closely approximate those of real classrooms (0.367–0.486), with only minor genre-specific variations. This demonstrates that EduVerse faithfully reproduces teacher-led discourse structures, ensuring authenticity at the interactional level.

\textbf{BEC(Behavior-Emotion-Cognition) across environments.}
Fig.~\ref{fig:students-behavior-distribution} shows that students’ states align with realistic classroom tendencies. Collaborative layouts foster the highest positive emotions (0.547) and higher-order cognition (0.261), round-table layouts exhibit more disengagement (0.254) and lower interaction (0.357), while lecture layouts mirror traditional classrooms, dominated by lower-order cognition (0.819) and passive or receptive behaviors. These results confirm that EduVerse effectively captures environment-dependent classroom dynamics.

\begin{figure}[t!]
    \centering
    \includegraphics[width=\textwidth]{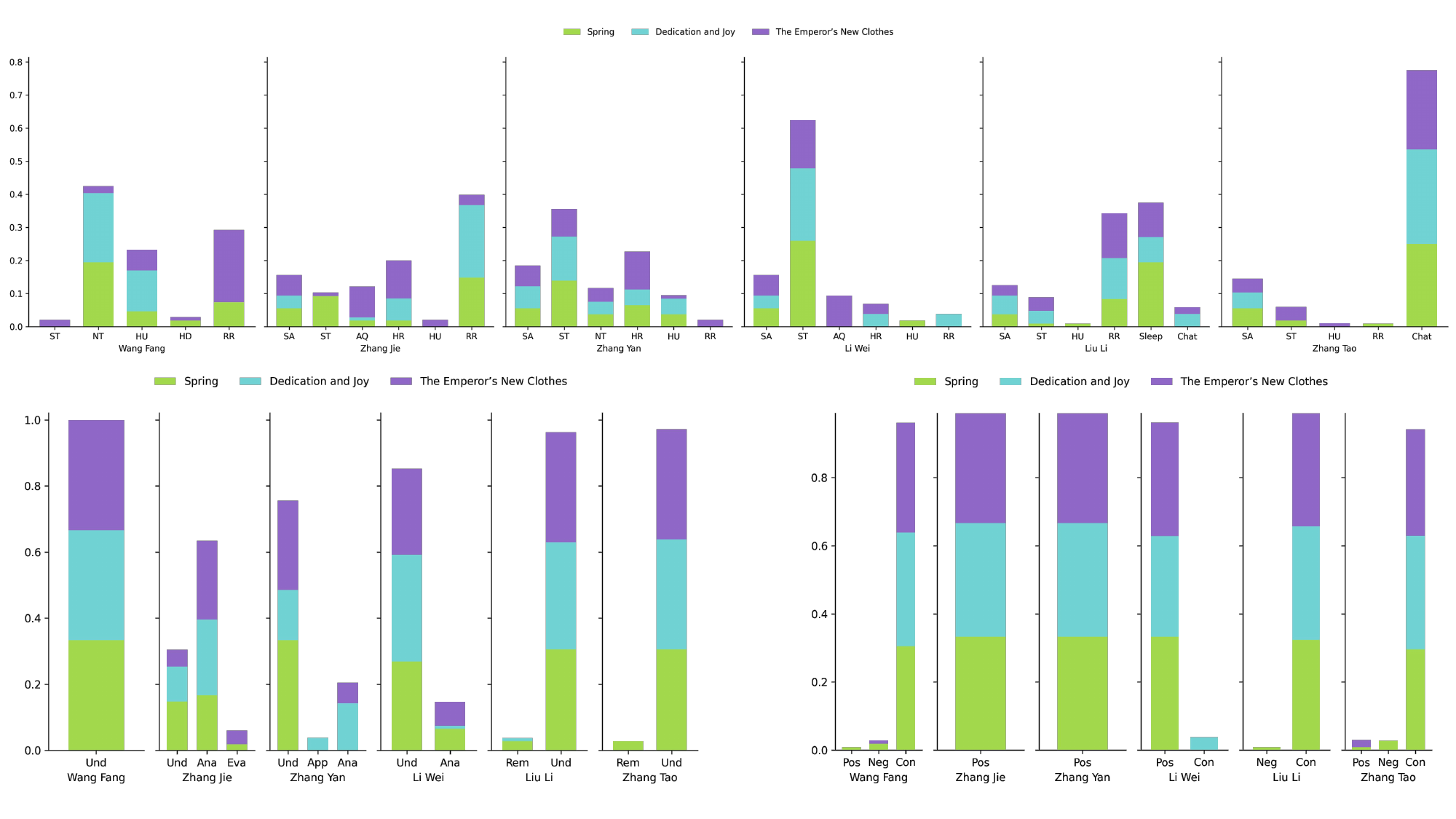}
    \caption{\textbf{Stable BEC patterns across genres within individual students.} 
Despite genre differences, individual patterns remain stable and align with traits—for instance, high-extraverted students show active engagement and varied cognition, while low-openness or low-conscientiousness students tend toward disengagement, lower-order cognition, and confusion.
}
    \label{fig:individual_bec}
\end{figure}

\textbf{Personality-driven stability.}
At the individual level, Fig.~\ref{fig:individual_bec} shows that students maintain stable BEC distributions across genres, aligned with their traits. For instance, \ZhangJie (high-extraversion) remains active and positive with diverse cognition, whereas \LiuLi (low-openness) and \ZhangTao (low-conscientiousness) exhibit disengagement, lower-order cognition, and frequent confusion. These results demonstrate EduVerse’s ability to preserve individual consistency and personality-conditioned behaviors.

\textbf{Ablation study.}
Fig.~\ref{fig:Ablation_study} further validates the necessity of key modules. Removing the stylization module results in evenly distributed emotions with amplified negativity, deviating from real classrooms where positive and confused states dominate. Removing the CIE module exaggerates higher-order cognition and active behaviors, resembling expert reasoning rather than gradual student development. Together, these results show that stylization ensures realistic emotional patterns, while the CIE module enforces educationally consistent cognitive and behavioral decisions, and their integration is indispensable for reproducing authentic classroom dynamics.

\textbf{Summary.}
Experiment I demonstrates that EduVerse reproduces authentic discourse, captures environment effects, preserves personality stability, and validates key modules, providing a solid basis for studying classroom design and instructional modes.

\subsection{Experiment II: Agent Customization for Group Interaction Analysis and Role Differentiation}
\label{subsec:exp-i}

\begin{wraptable}{r}{0.6\textwidth}
  \centering
  \vspace{-10pt}
  \caption{Group interaction analysis across lessons and environments. 
  Values report nodes, edges, density, and average degree of the interaction graph. 
  \textbf{Bold} marks the highest values per lesson.}
  \label{tab:group-interaction}
  \vspace{-5pt}
  \setlength{\tabcolsep}{4pt}
  \small
  \begin{tabular}{p{1.8cm}lcccc}
    \toprule
    \textbf{Lesson} & \textbf{Env.} & \textbf{Nodes} & \textbf{Edges} & \textbf{Density} & \textbf{Avg.~Deg.} \\
    \midrule
    \multirow{3}{*}{\makecell[l]{\textit{Foreign} \\ \textit{Fiction}}}      
      & \textit{Lecture} & 6 & 5 & \textbf{0.333} & \textbf{1.667} \\
      & \textit{Collab}  & 5 & 3 & 0.300          & 1.200 \\
      & \textit{Round}   & 6 & 5 & \textbf{0.333} & \textbf{1.667} \\
    \midrule
    \multirow{3}{*}{\makecell[l]{\textit{Argumentative} \\ \textit{Essay}}} 
      & \textit{Lecture} & 5 & 3 & 0.300          & 1.200 \\
      & \textit{Collab}  & 5 & 4 & \textbf{0.400} & \textbf{1.600} \\
      & \textit{Round}   & 6 & 5 & \textbf{0.333} & \textbf{1.667} \\
    \midrule
    \multirow{3}{*}{\makecell[l]{\textit{Lyrical} \\ \textit{Prose}}}        
      & \textit{Lecture} & 6 & 4 & 0.267          & 1.333 \\
      & \textit{Collab}  & 5 & 3 & 0.300          & 1.200 \\
      & \textit{Round}   & 6 & 5 & \textbf{0.333} & \textbf{1.667} \\
    \bottomrule
  \end{tabular}
  \vspace{-10pt}
\end{wraptable}

This experiment evaluates EduVerse’s ability to model interaction, moving from group-level networks to individual influence, and finally testing human–agent integration via the open interface.

\begin{wraptable}{r}{0.5\textwidth}
  \centering
  \caption{Distribution of students’ network centrality indicators in the \textit{Dedication and Joy} lesson across environments.
  Values are normalized to $[0,1]$; \textbf{Deg.} denotes degree centrality and \textbf{Betw.} denotes betweenness centrality.}
  \label{tab:centrality_dj}
  \setlength{\tabcolsep}{4pt}
  \small
  \begin{tabular}{p{1cm}lcccc}
    \toprule
    \textbf{Env.} & \textbf{Student} & \textbf{In} & \textbf{Out} & \textbf{Deg.} & \textbf{Betw.} \\
    \midrule
    \multirow{5}{*}{\textit{Lecture}}
      & \LiWei    & 0.25 & 0.25 & 0.50 & 0    \\
      & \LiuLi    & 0.25 & 0.25 & 0.50 & 0    \\
      & \ZhangTao & 0.00 & 0.25 & 0.25 & 0    \\
      & \ZhangJie & 0.50 & 0.00 & 0.50 & 0    \\
      & \ZhangYan & 0.00 & 0.25 & 0.25 & 0    \\
    \midrule
    \multirow{5}{*}{\textit{Collab}}
      & \LiWei    & 0.25 & 0.25 & 0.50 & 0     \\
      & \LiuLi    & 0.25 & 0.25 & 0.50 & 0     \\
      & \ZhangTao & 0.50 & 0.25 & 0.75 & 0.083 \\
      & \ZhangYan & 0.25 & 0.50 & 0.75 & 0.083 \\
      & \ZhangJie & 0.25 & 0.25 & 0.50 & 0     \\
    \midrule
    \multirow{6}{*}{\textit{Round}}
      & \LiWei    & 0.20 & 0.20 & 0.40 & 0    \\
      & \ZhangJie & 0.40 & 0.20 & 0.60 & 0.10 \\
      & \LiuLi    & 0.00 & 0.20 & 0.20 & 0    \\
      & \WangFang & 0.40 & 0.00 & 0.40 & 0    \\
      & \ZhangTao & 0.20 & 0.60 & 0.80 & 0.15 \\
      & \ZhangYan & 0.20 & 0.20 & 0.40 & 0    \\
    \bottomrule
  \end{tabular}
\end{wraptable}

\textbf{Group Interaction Analysis.}
We model classroom interactions as undirected graphs, using density ($D=\tfrac{2E}{N(N-1)}$) and average degree ($k=\tfrac{2E}{N}$). As shown in Tab.~\ref{tab:group-interaction}, density (0.267–0.400) and degree (1.2–1.667) indicate that 27–40\% of ties are realized, reflecting realistic yet localized participation in teacher-led classrooms. Genre–environment effects also appear: argumentative essays and prose show lower density in Lectures, while fiction remains around 0.3 across settings, underscoring the narrative appeal of storytelling. Overall, genre and layout jointly shape engagement.

\textbf{Individual Influence Analysis.}
We computed four directed metrics—\textit{in-degree} (attention), \textit{out-degree} (initiative), \textit{degree centrality} (activity), and \textit{betweenness} (bridging). As shown in Tab.~\ref{tab:centrality_dj}, the \textit{Dedication and Joy} lesson reveals mode-dependent roles. In Lecture, \ZhangJie had high in-degree but no initiative, while \ZhangTao and \ZhangYan showed the opposite. In Collab, reciprocity increased, with \ZhangTao emerging as a connector and \ZhangJie shifting to side-talk. In Round, roles decentralized, yet \ZhangTao became most central (highest out-degree, degree, and betweenness), whereas \LiuLi was marginalized. (see Fig.~\ref{fig:student-interaction} for the visualization of student interactions). These results illustrate how classroom organization reshapes roles, moving individuals from peripheral to core positions.

\textbf{Human–Agent Interaction.}
To test human integration, we designed four subtasks: peer chatting, peer academic response, teacher Q\&A, and teacher intervention (Fig.~\ref{fig:rader}). Results aligned with personality traits: \ZhangTao (talkative) responded most (0.53–0.73) and initiated conversations in Collab/Round, while \ZhangJie (high-extraversion) responded less (0.27–0.40). Teacher agents succeeded in all Q\&A and interventions, ensuring robust instructional control. These findings confirm EduVerse’s capacity for seamless human integration with role-driven realism.

\textbf{Summary.}
EduVerse captures group patterns, individual traits, and human–agent integration, validating its ability to model authentic and adaptive classroom interactions.

\subsection{Experiment III: Session Customization for Cross-Session Evolution and Long-term Development}
\label{subsec:exp-e}

\begin{wrapfigure}{r}{0.5\textwidth}
\vspace{-10pt}
\centering
\includegraphics[width=0.5\textwidth]{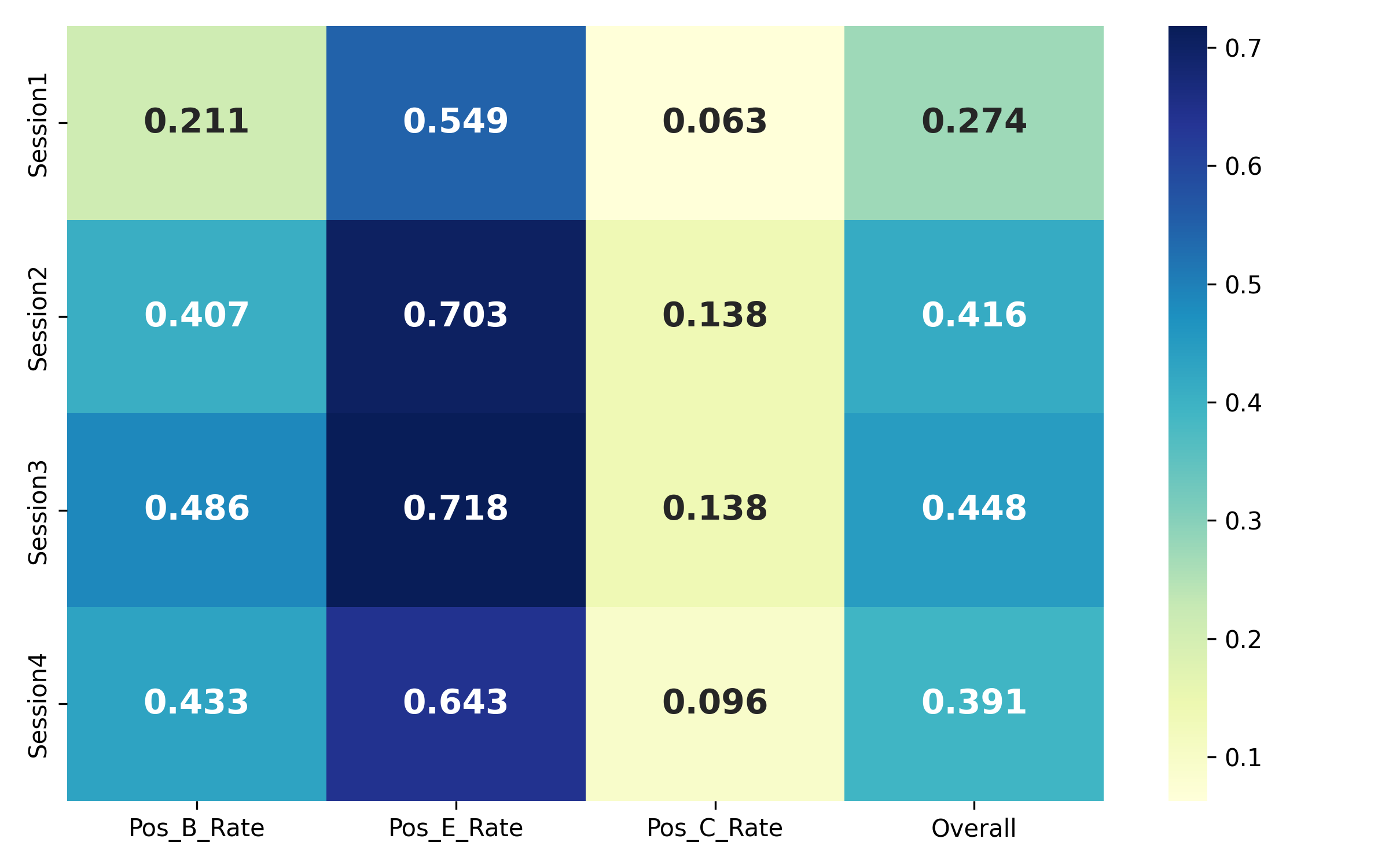}
\caption{\textbf{Positive transition trends in cross-session evolution.} Rates of positive shifts in behavior, emotion, and cognition increase over time, with behavior improving most rapidly, emotion rising steadily, and cognition progressing gradually.}
\label{fig:positive-transition}
\end{wrapfigure}

This experiment examines whether virtual students show progressive development across extended instruction by mapping BEC to ordered levels and defining upward transitions as positive shifts. Tracking four sessions (\textit{Spring} I–II, \textit{Dedication and Joy} I–II), we evaluate long-term engagement and learning trajectories.

\textbf{Session-level Evolution.}
As shown in Fig.~\ref{fig:positive-transition}, positive shifts increase across sessions. Behavior improves fastest, moving from passive to interactive under teacher scaffolding. Emotions remain high and rise early but plateau with the demands of argumentative texts. Cognition advances more slowly, reflecting the need for sustained accumulation. These trends show that EduVerse captures authentic dynamics where behavior and affect change quickly, while cognition develops gradually.

\begin{figure}[t!]
\centering
\includegraphics[width=\textwidth]{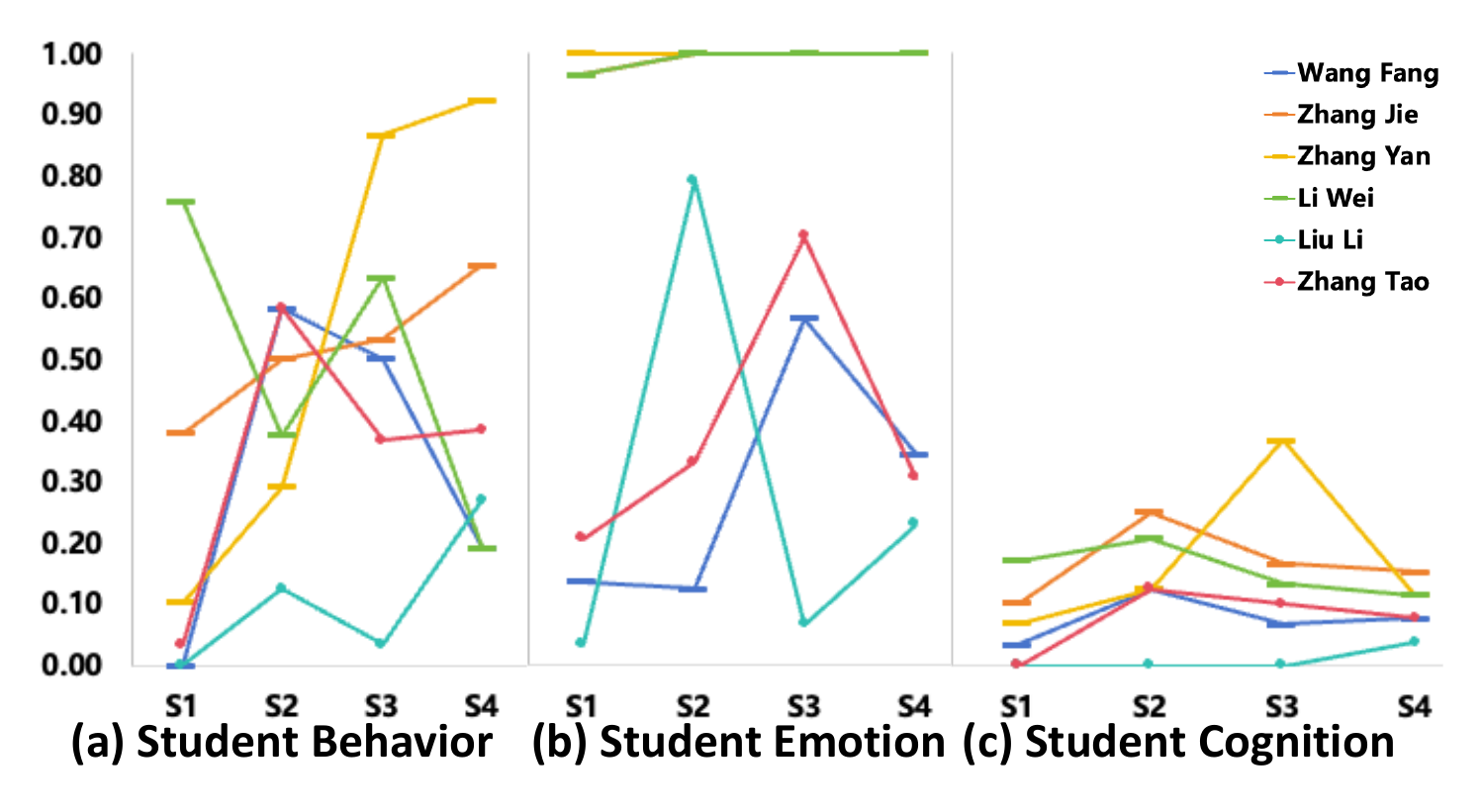}
\caption{\textbf{Individual trajectories of positive shifts across sessions.} Students display differentiated developmental paths in behavior, emotion, and cognition, closely aligned with personality traits, validating EduVerse’s capacity to model personalized learning evolution.}
\label{fig:three_images}
\end{figure}

\textbf{Individual-level Evolution.}
Fig.~\ref{fig:three_images} shows six student trajectories with clear individual differences. \WangFang improved steadily with moderate adaptation; \ZhangJie advanced consistently across all dimensions, reflecting high extraversion; \ZhangYan made a major cognitive leap in session three, showing adaptability; \LiWei sustained positive affect but lagged in learning; \LiuLi started low but improved gradually, typical of low-openness; \ZhangTao was volatile in behavior and emotion with limited cognitive gains. Overall, EduVerse captures realistic, individualized learning pathways.

\textbf{Summary.}
The results demonstrate the long-term evolutionary value of EduVerse: behavior and emotion show rapid short-term improvements, while cognition develops more gradually. Moreover, individual trajectories diverge according to personality traits and learning styles, underscoring the realism and applicability of the framework for simulating educational evolution.

\subsection{Summary}
Our experiments collectively demonstrate the robustness and applicability of the EduVerse framework as a simulation space for educational research. Experiment I validates classroom authenticity and personality stability, confirming realistic behavioral patterns in customized environments. Experiment II captures group-level dynamics and supports seamless human–agent collaboration, highlighting both structural coherence and adaptive flexibility. Experiment III traces long-term behavioral, emotional, and cognitive trajectories, revealing progressive gains alongside individual divergences driven by personality traits. Together, these results establish EduVerse as a reliable paradigm for reproducing key classroom features and provide strong evidence of its value for studying learning processes and supporting teacher training.

\section{Conclusion}
\label{sec:conclusion}

We introduced \textbf{EduVerse}, the first user-defined multi-agent simulation space for education that unifies cognition, interaction, and evolution (CIE) within a single framework. EduVerse enables trait-conditioned, metacognitive student agents that exhibit consistent yet adaptive behaviors across diverse instructional contexts.
Validated in middle school Chinese language classrooms, EduVerse demonstrates robust alignment with real-world dynamics. Experiment I confirms that customized environments produce realistic learning patterns and stable personality-driven behaviors, achieving an average IRF completion rate of 42.28\%. Experiment II shows that group-level interaction density ranges from 0.27–0.40—about one-third of potential peer connections—while also enabling seamless human–agent integration. Experiment III traces longitudinal development, with overall positive transitions increasing by 11.7\%, though individual trajectories diverge according to personality traits and learning styles.
These findings position EduVerse as a scalable, interpretable, and transferable framework for modeling developmental student agents. By bridging simulation with authentic educational processes, EduVerse provides a methodological foundation for future research in adaptive learning systems, human-in-the-loop pedagogy, and cross-disciplinary applications of educational AI.


\section*{Ethics Statement}
This work follows ethical standards and data usage regulations. EduVerse is built from publicly available classroom materials, with all content anonymized to remove identifiable information. Student agents are modeled only with abstract personality traits and interaction patterns, without including demographic or sensitive attributes. Human participation in the experiments was limited to controlled simulations, with no private data collected.

\section*{Reproducibility Statement}
To support reproducibility, we will release the EduVerse codebase, environment settings, preprocessing scripts, and trained models. Processed resources and evaluation protocols will also be provided. All resources will be made available upon acceptance.

\color{black}

\bibliography{iclr2026_conference}
\bibliographystyle{iclr2026_conference}

\clearpage

\appendix

\renewcommand{\thetable}{A\arabic{table}}
\renewcommand{\thefigure}{A\arabic{figure}}
\renewcommand{\thealgocf}{A.\arabic{algocf}}  
\setcounter{figure}{0}  
\setcounter{table}{0}  
\setcounter{algocf}{0} 

\section*{Appendix}


\section{Comprehensive Related Works}
\label{sec:more-relate-work}

\subsection{Educational Agents and Virtual Student Simulations}
\label{subsec:related-work-1}

Modeling virtual student agents has long been a central thread in educational AI research~\citep{dai2022educational,chou2003redefining,chheang2024towards}. As early as the 1990s, researchers introduced ``teachable agents'' to support teacher training and foster learning by teaching. Representative examples include VanLehn et al.’s work in physics tutoring~\citep{vanlehn1994applications} and the Betty’s Brain system~\citep{biswas2005learning}, which formalized the learning-by-teaching paradigm. However, these systems relied heavily on scripted behavioral policies and static knowledge-update mechanisms, limiting their ability to capture temporal dynamics in student behavior and affect. Similarly, traditional intelligent tutoring systems emphasized knowledge mastery (e.g., skill tracing) but paid little attention to developmental trajectories.

The advent of large language models (LLMs) has enabled a new generation of educational agents, characterized by open-ended interaction, behavioral diversity, and learner-like imperfections~\citep{russin2024frege,chu2025llm,moore2023empowering}. Recent work has leveraged prompt-conditioned LLMs to simulate students with varying ability levels, personalities, and misconceptions~\citep{wang2024can,glaese2022improving}. For example, Lu and Wang~\citep{lu2024generative} proposed the Generative Students framework to assess item difficulty through simulated learners; Markel et al.~\citep{markel2023gpteach} introduced GPTeach to enhance teacher training with diverse LLM-generated student responses; and Jin et al.~\citep{jin2025teachtune} developed TeachTune, which evaluates instructional agents against a spectrum of student personas, highlighting how pedagogical strategies adapt to personality-driven differences. Building on these advances, EduPersona~\citep{zhu2025edupersona} systematically benchmarks virtual students’ subjective ability boundaries across personality types, while the SOEI framework~\citep{ma2024llms} introduces structured teacher–student interaction modeling and multi-dimensional evaluation mechanisms, laying the foundation for personality-consistent and behaviorally realistic student agents.

Beyond individual interactions, researchers have explored classroom-level simulations with multi-agent LLM frameworks. AgentVerse~\citep{chen2023agentverse} and CGMI~\citep{jinxin2023cgmi} enable heterogeneous agents to engage in collaborative and instructional roles. Building on this, SimClass~\citep{zhang2024simulating} incorporates a class manager to coordinate Initiation–Response–Feedback (IRF) dialogues between teacher and student agents, achieving high realism: agents respond to teacher prompts, initiate follow-ups, maintain turn-taking, and even exhibit emergent behaviors such as spontaneous group discussions and peer-led task completion. In parallel, SOEI~\citep{ma2024llms} provides a personality-modulated framework for structured classroom interactions, enhancing the realism and interpretability of classroom simulations from both interactional and behavioral perspectives—marking a shift from scripted interactions toward socially grounded, self-organizing learning environments.

Nevertheless, as recent review studies have summarized~\citep{tan2025artificial,tan2025comprehensive}, existing systems remain limited in modeling long-term behavioral evolution, affective regulation, and stylistic coherence, with most constrained to single-session interactions. In contrast, \textbf{EduVerse} provides a unified, multi-level simulation platform that supports user-defined modeling of physical environments, agent configurations, and cross-session evolution. Its open interfaces allow researchers to flexibly configure classroom layouts, seating arrangements, and interaction networks, and systematically examine learners’ cognitive development and group dynamics. Through this design, EduVerse bridges the gap between scripted tutor–learner simulations and scalable, development-oriented classroom modeling.

\subsection{LLM-driven Multi-agent Social Simulations}
\label{subsec:related-work-2}

The rise of large language models (LLMs) has fueled multi-agent social simulations in which agents act as generative entities capable of open-ended dialogue and socially coherent behavior~\citep{guo2024large,gao2024large,hua2023war}. A landmark example, \textit{Generative Agents}~\citep{park2023generative}, deployed 25 GPT-powered agents in a virtual town: agents followed daily routines, formed human-like relationships, and even organized a collective event without external control, illustrating emergent behavior grounded in persistent memory and reflection. Follow-up studies extended this paradigm to collaborative and adversarial settings: ChatDev~\citep{qian2023chatdev} and MetaGPT~\citep{hong2023metagpt} simulate professional teams via role-specific prompting and structured communication flows; Wang et al.~\citep{wang2023unleashing} emulate multi-agent reasoning through self-dialogue. In competitive contexts, Xu et al.~\citep{xu2023exploring} model deception, persuasion, and coalition in Werewolf, while Horton~\citep{horton2023large} and Aher et al.~\citep{aher2023using} study negotiation and public-goods games, revealing rational yet human-biased behaviors.

Research in dynamic vision offers methodological insights that are directly relevant here: model capability is highly sensitive to non-stationary, topology-shifting environments, and thus capability measurement should be co-designed with environmental generation and evaluation protocols. One line of work adopts a 3E paradigm (Environment–Evaluation–Executor) to build user-defined task spaces, using per-frame challenge factors and new indicators to expose systematic performance differences of the same model under distinct environmental configurations~\citep{hu2024sotverse}. Other efforts emphasize a triad of a challenging task, a fair environment, and a scientific evaluation procedure, with human performance as a reference~\citep{hu2023global,hu2024multi,zhao2024biodrone,wang2025soi,zhang2025beyond}. These observations align with evaluation perspectives that treat environment, data, and evaluation as co-equal factors for characterizing capability~\citep{zhao2025visual,hu2024fiova}.

From an educational standpoint, most existing LLM-based social simulations focus on adult, informal contexts and typically lack structured instructional discourse, long-horizon learner modeling, and explicit treatment of knowledge progression and group dynamics. \textbf{EduVerse} introduces LLM-driven multi-agent simulation into education with user-definable physical and social environments and integrates structured classroom discourse (e.g., IRF cycles) with cross-session evolution, enabling interpretable analyses of instructional interactions and learner development under explicitly controlled dynamic factors.

\subsection{Personalized Modeling and Behavioral Consistency Mechanisms}
\label{subsec:related-work-3}

Maintaining long-term consistency of personality, language style, and behavioral patterns remains a central challenge in LLM-driven multi-agent environments~\citep{guo2024large,li2024survey,tran2025multi}. Despite their generative fluency, LLMs often suffer from “persona drift” in extended interactions, where role-specific traits weaken or deviate over time~\citep{xu2023exploring}. For example, a student agent initially designed to be introverted may gradually adopt assertive conversational patterns, undermining the credibility of long-term simulations. Such issues arise from limited identity retention and emotional coherence, motivating research on role stabilization mechanisms.

Early dialogue systems attempted to preserve character identity through structured profiles and task-specific memory modules. PersonaChat~\citep{zhang2018personalizing}, for instance, embedded fixed persona facts, while later work introduced memory modules to retain role-consistent traits across turns~\citep{ouyang2022training}. These approaches proved effective in short sessions but struggled with complexity and duration. In LLM-based agents, persona-conditioned prompting became a common strategy, though it often failed under topic shifts or multi-phase tasks~\citep{xu2023exploring}.

Recent advances emphasize fine-tuning and reward-based alignment to improve role fidelity~\citep{ji2025survey,chen2025step}. CharacterGLM~\citep{zhou2023characterglm} and Ditto~\citep{lu2024large} enhanced intra-role consistency by training on persona-labeled dialogues or generating synthetic role-specific corpora. Reinforcement Learning from Human Feedback (RLHF) further penalized off-character outputs, reinforcing behavioral alignment. However, these methods face scalability challenges, including data costs and generalization trade-offs.

To mitigate reliance on supervised correction, self-monitoring strategies have emerged~\citep{li2025ai,behore2024enhancing}. Ji et al.~\citep{ji2025enhancing} introduced a role-aware reflection loop that allows models to detect and revise misaligned outputs. Coupled with contrastive training, this approach significantly improved long-term role consistency. Memory-augmented designs provide additional scaffolding: systems log key interactions and behavioral states, with summarization or vector retrieval supporting continuity in future outputs~\citep{park2023generative,ouyang2022training}.

These developments are particularly salient in education. To simulate realistic learners, researchers have designed student agents with distinct cognitive and non-cognitive traits. TeachTune~\citep{jin2025teachtune} showed that teacher agents adjust feedback strategies depending on student profiles such as confidence or anxiety, while Li et al.~\citep{li2025exploring} modeled “imperfect learners” by injecting errors into outputs, prompting teacher agents to practice remediation. Such findings highlight the pedagogical utility of stable yet differentiated learner personas.

\textbf{EduVerse} advances this line of work by integrating profile-driven planning, memory-aligned regulation, and behavior-consistent generation into its Cognition–Interaction–Evolution framework. Compared to single-agent approaches, EduVerse achieves both intra-agent coherence and inter-agent variation at the classroom scale, providing a scalable and interpretable solution for educational simulation.

\clearpage

\section{Detailed Information for EduVerse Framework}
\label{app:more-EduVerse-framework}

\subsection{Detailed Description of Environment Module}

\label{subapp:environment}

\subsubsection{Hierarchical Spatial Organization}

EduVerse models classroom environments using a four-tiered hierarchical spatial structure $\mathcal{Z} = \{\mathcal{Z}_S,\mathcal{Z}_A,\mathcal{Z}_O,\mathcal{Z}_I\}$, corresponding to \textbf{Sector}, \textbf{Arena}, \textbf{Object}, and \textbf{Item}. This design supports pedagogically meaningful interactions and dynamic behavior generation, and provides semantic alignment between agent perception, spatial context, and task execution.

At the top level, a \textbf{Sector} partitions the virtual classroom into functional instructional zones, e.g., \textit{Teacher Sector}, \textit{Student Sector}, and \textit{Activity Sector}. Within each sector, an \textbf{Arena} specifies localized interactive regions—such as a discussion circle or a teaching podium—that constrain mobility and determine perceptual access to nearby objects. \textbf{Objects} embedded within Arenas (e.g., desks, blackboards, presentation screens) serve as anchors for attention and instructional actions. \textbf{Items} (e.g., chalk, textbooks, notebooks) are the most granular perceivable units and constitute the basic elements of fine-grained interaction.

For example, in the \textit{Teacher Podium Arena}, a blackboard and lectern support behaviors such as lecturing and board writing. In contrast, a \textit{Student Group Arena} contains desks and personal learning materials, enabling small-group collaboration or individualized learning. A complete layout instance typically includes multiple sectors, each containing several arenas populated with pedagogically structured objects and items (e.g., chalk, erasers, water cups), collectively forming a coherent four-level spatial graph (see Tab.~\ref{tab:env-structure}). This hierarchical structure supports classroom configurations ranging from lecture-centric layouts to collaborative or round-table setups, and allows real-time adaptation of spatial roles, interaction boundaries, and perception zones as classroom dynamics evolve.

\begin{table}[htbp]
\centering
\setlength{\tabcolsep}{2pt}
\small
\begin{threeparttable}
\caption{\textbf{Hierarchical Classroom Environment Structure}}
\label{tab:env-structure}
\vspace{0.5em}
\begin{tabular}{>{\raggedright\arraybackslash}p{9em} >{\raggedright\arraybackslash}p{9.5em} >{\raggedright\arraybackslash}p{8em} >{\raggedright\arraybackslash}p{15em}}
\toprule
\textbf{Sector (ID)} & \textbf{Arena} & \textbf{Object} & \textbf{Items} \\
\midrule
Teacher Zone (10001) & Resource Display Area & Blackboard & Chalk, Blackboard Eraser \\
Teacher Zone (10001) & Podium Area & Podium Desk & Cup, Textbook, Mobile Phone, Chalk \\
Teacher Zone (10001) & Instruction Area & Electronic Whiteboard & — \\
\midrule
Student Zone (20001) & Group A Area & Student $A_{1\ldots N}$ & Textbook, Cup, Pencil Case, Pen, Backpack \\
Student Zone (20001) & Group B Area & Student $B_{1\ldots N}$ & — \\
\midrule
Activity Zone (30001) & Storage Area & Cleaning Supplies & Broom, Mop \\
Activity Zone (30001) & Storage Area & Daily Utilities & Clock, Cabinet, Water Dispenser \\
\bottomrule
\end{tabular}
\vspace{0.5em}
\begin{tablenotes}
\footnotesize
\item[] Note: This table instantiates the 4-level spatial structure in \textbf{EduVerse}—Sector, Arena, Object, and Item. Each arena defines a functional subspace containing interactive objects for perception and behavior planning.
\end{tablenotes}
\end{threeparttable}
\end{table}

\subsubsection{Interaction Affordances and Constraints}

The spatial hierarchy also functions as an interaction scaffold that governs how agents perceive, move, and act. Layouts map to pedagogical strategies: lecture-centric configurations emphasize teacher-led, unidirectional communication, while round-table or collab-based layouts promote peer collaboration. The ``Sector–Arena–Object–Item'' hierarchy accommodates these variations and enables dynamic spatial adaptation during simulation.

In the current implementation, spatial location primarily conditions behavioral availability. For example, only students within a \textit{Presentation Arena} can access teacher content in real time; in group tasks, shared Arena–Object associations trigger peer-based interactions. EduVerse thus enforces mode-specific constraints (lecture, presentation, group discussion), with corresponding action permissions. 

Beyond this, the framework is designed to be extensible. For instance, each student can be assigned a personalized \texttt{perception\_config} (e.g., perceptual radius and maximum trackable items), and collision constraints or spawn configurations can be integrated to enhance physical realism. While not activated in the current experiments, these extensions illustrate the scalability of EduVerse towards more fine-grained behavioral planning and environment fidelity.

\subsubsection{Seat\_Graph Formalization and Layout Templates}
We define the seat graph as an unweighted adjacency matrix 
$A^{\text{seat}} \in \{0,1\}^{N \times N}$, where
\[
A^{\text{seat}}_{ij} =
\begin{cases}
1, & \text{if $i$ and $j$ satisfy the layout-specific adjacency rule}, \\
0, & \text{otherwise}.
\end{cases}
\]

Formally, EduVerse provides three canonical layouts:

\begin{itemize}
    \item Lecture
Adjacency is primarily defined within rows, restricted to students sitting 
close to each other:
\[
A^{\text{seat}}_{ij} = \mathbb{I}\big[d(i,j) \leq \tau_d \ \wedge \ \text{row}(i)=\text{row}(j)\big].
\]

    \item Round Table
Students are seated around circular tables; adjacency includes both immediate 
neighbors and face-to-face counterparts:
\[
A^{\text{seat}}_{ij} = \mathbb{I}\big[j=\mathrm{neighbor}(i) \ \lor \ j=\mathrm{opp}(i)\big].
\]

    \item Collab\_Two Tables
Within-group students are fully connected, while across-group edges are suppressed:
\[
A^{\text{seat}}_{ij} = \mathbb{I}\big[g(i)=g(j)\big].
\]
\end{itemize}

Thus, the seat graph provides a baseline, layout-dependent topological structure:
\[
A^{\text{seat}} =
\begin{cases}
A^{\text{seat}}_{\text{Lecture}}, & \text{if layout = Lecture}, \\[4pt]
A^{\text{seat}}_{\text{Round}},   & \text{if layout = Round Table}, \\[4pt]
A^{\text{seat}}_{\text{Collab}},  & \text{if layout = Collab\_Two\_Tables}.
\end{cases}
\]

This structure is a binary, unweighted topology that captures only peer adjacency. It does not yet encode higher-order social or instructional factors. Instead, it serves as a foundation upon which \emph{cognitive planning} can later be modeled by incorporating teacher and board visibility.

\subsubsection{Cognitive Planning and Distance Effects}

The physical distances encoded in the seat graph directly affect students' cognitive planning. 
While the seat graph itself captures only peer proximity relations, the cognitive planning process 
requires a richer consideration of spatial constraints, including student--student, student--teacher, 
and student--board distances. Importantly, these spatial factors constitute only \emph{one component} 
of cognitive planning; other psychological, pedagogical, and contextual variables also play a crucial role, 
but here we explicitly highlight the influence of the physical environment.

\textbf{Distance-based cognitive factors}

Let $p_i \in \mathbb{Z}^2$ denote the spatial position of student $i$. We define three forms of 
distance measures relevant to cognitive planning:

\begin{itemize}
    \item \textbf{Peer distance:}
    \[
    d(i,j) = \|p_i - p_j\|_2, \quad i \neq j,
    \]
    capturing the Euclidean distance between students $i$ and $j$.
    
    \item \textbf{Teacher distance:}
    \[
    d_T(i) = \|p_i - p_T\|_2,
    \]
    where $p_T$ denotes the teacher's position. This distance reflects the cognitive accessibility 
    of the teacher, which is essential for attention allocation and interaction.
    
    \item \textbf{Board distance:}
    \[
    d_B(i) = \|p_i - p_B\|_2,
    \]
    where $p_B$ denotes the board's position. This measure captures the visibility and salience 
    of instructional materials.
\end{itemize}

\textbf{Integration into cognitive planning}

In cognitive planning, a student's effective engagement is influenced by a combination of these 
distances. We define a cognitive planning function
\[
C(i) = f\big(\{d(i,j)\}_{j \neq i}, \ d_T(i), \ d_B(i), \ \Omega(i)\big),
\]
where $f(\cdot)$ aggregates peer proximity, teacher distance, and board distance together with 
additional factors $\Omega(i)$ (e.g., individual motivation, prior knowledge, or task demands). 

Thus, physical proximity among peers (as encoded in the seat graph) provides the structural 
baseline, while teacher and board distances add instructional and attentional dimensions. 
These spatial factors, combined with non-spatial determinants $\Omega(i)$, jointly shape 
the student's cognitive planning process.

\subsubsection{Configuration Schema and Physical Realism}
Our environment is instantiated on a two-dimensional grid of size 
$30 \times 20$. Each classroom layout is parameterized by a configuration schema, 
including layout type (\textit{Lecture}, \textit{Round Table}, or \textit{Collaborative Two Tables}), 
table radius, table spacing, and object coordinates (students, teacher, podium, board, and tables). 

Peer adjacency is determined by a Euclidean distance threshold $\tau_d = 4.5$, 
with an additional tolerance of $+1.0$ if two students belong to the same group. 
In the \textit{Round Table} layout, opposite students are explicitly connected, 
while in the \textit{Collaborative} layout, within-table students form a complete subgraph 
and inter-table connections are suppressed. 

Perception is bounded by a radius ($r=8$) and a capacity constraint ($\kappa=5$), 
ensuring that each student can only attend to a limited number of peers or objects. 
Collision masks and occupancy constraints are implicitly enforced by the grid representation, 
while actions violating spatial feasibility (e.g., moving into occupied cells or colliding 
with fixed objects) are rejected. This ensures reproducibility and interpretability 
of spatially grounded interactions.

\begin{table}
  \centering
  \caption{Environment configuration parameters in EduVerse. 
  Parameters specify spatial layout, interaction rules, and perception constraints.}
  \label{tab:env_config}
  \setlength{\tabcolsep}{4pt}
  \small
  \begin{tabular}{ll}
    \toprule
    \textbf{Parameter} & \textbf{Description} \\
    \midrule
    Grid size          & $30 \times 20$ discrete lattice \\
    Layout type        & Lecture / Round Table / Collaborative (two tables) \\
    Table radius       & Default $=3$ grid units \\
    Table gap          & Default $=10$ grid units (between two tables) \\
    Objects            & Students, teacher, podium, board, tables \\
    Peer adjacency     & Edge $(i,j)$ if $d(i,j) \leq \tau_d = 4.5$ \\
    Group adjustment   & Threshold $+1.0$ if $g(i) = g(j)$ \\
    Round Table rule   & Add cross-table ``opposite'' edges \\
    Collaborative rule & Within-table complete graph; no inter-table edges \\
    Perception radius  & $r = 8$ grid units \\
    Perception capacity& $\kappa = 5$ objects per student \\
    Collision/occupancy& One object per grid cell; invalid moves rejected \\
    \bottomrule
  \end{tabular}
\end{table}

\subsection{Detailed Description of Cognition Engine}
\label{subapp:cie-c}

\subsubsection{Theoretical Foundations of PCA Architecture}

CIE adopts a three-stage \textbf{Perception–Cognition–Action (PCA)} architecture to simulate agent decision-making~\citep{chen2024pca,davis2003recognizing,hancock1996face,yao2020tri}. This design is grounded in both classical AI agent models and foundational theories in educational psychology, capturing the trajectory from environmental sensing to goal-directed behavior.

From a psychological perspective, PCA aligns with the Atkinson–Shiffrin model of information processing~\citep{atkinson1968human,izawa1999human,cheng2022xmem} and resonates with constructivist and metacognitive theories of learning~\citep{gunstone1992constructivism,bonanno2004metacognition}. It integrates affect, cognition, and behavior, emphasizing how learners perceive, interpret, and regulate actions in evolving classroom contexts.

Within this structure, the \textbf{Perception} stage gathers contextual cues such as teacher prompts, peer behaviors, and environmental signals. The \textbf{Cognition} stage transforms these inputs into internal reasoning and action intent, and the \textbf{Action} stage executes verbal, physical, or social behaviors accordingly.

A central innovation in CIE is the integration of a metacognitive loop—\textbf{Plan, Monitor, Regulate}—within the cognition stage. Student agents are not merely reactive: they proactively plan whether and how to act, monitor the clarity and outcomes of their behaviors, and regulate subsequent strategies based on feedback. For example, an agent may decide whether to speak (\textbf{Plan}), evaluate the appropriateness of its utterance (\textbf{Monitor}), and adjust behavior after receiving teacher feedback (\textbf{Regulate}). This bidirectional cycle links teacher interventions with student adaptations, producing temporally grounded learning trajectories that approximate authentic development.

The same PCA structure extends to teacher agents, where cognitive stages underpin pedagogical decisions such as goal-setting, feedback selection, interpretation of class-level signals, and adjustment of emotional tone. This unified yet role-differentiated framework enables interpretable and traceable behaviors across heterogeneous agents.

In summary, the PCA architecture provides a theoretically grounded, modular, and extensible control mechanism, serving as the cognitive backbone of the CIE framework.

\subsubsection{Personality-conditioned Cognition}

\begin{table}[t!]
\small
\setlength{\tabcolsep}{2pt}
\centering
\caption{Functional differentiation between student and teacher agents in \textbf{CIE}.}
\label{tab:agent-diff}
\begin{tabularx}{\linewidth}{lXX}
\toprule
\textbf{Component} & \textbf{Student Agent $\mathcal{A}_S^i$} & \textbf{Teacher Agent $\mathcal{A}_T$} \\
\midrule
Perception Input & Teacher actions, peer behavior, social requests, seat graph, interaction graph & Class-wide metrics, participation logs \\
Cognitive Objective & Interpret prompts, decide on participation & Track engagement, guide task flow \\
Action Output & Interactive action, Active action, Passive action, Off-task action & Task instruction, Interactive teaching acts \\
Triggering Condition & Willingness + gating check & Scripted plans + real-time updates \\
Prompt Conditioning & \textit{Personalized traits:} personality, style, motivation & \textit{Fixed templates:} instructional role \\
\bottomrule
\end{tabularx}
\end{table}

To capture individualized learning dynamics, \textbf{CIE} embeds a triadic psychological model—\textbf{personality traits, cognitive style, and learning motivation}—into each student agent’s cognition engine. This design ensures that agents are not only responsive to instructional context but also conditioned by stable psychological dispositions, thereby enabling heterogeneous patterns of planning, action execution, and self-regulation.

This modeling approach is anchored in established educational psychology theories.  
(1) The \textit{Big Five Personality Traits} inform dispositional tendencies such as agreeableness or conscientiousness~\citep{gerber2011big,komarraju2011big}.  
(2) Dimensions like \textit{Field Dependence–Independence} and \textit{Impulsivity–Reflection} characterize cognitive style, shaping how learners process information and regulate decisions~\citep{swinnen1986role,jamieson1992cognitive,yang2023investigating}.  
(3) Motivational dimensions are guided by \textit{Self-Determination Theory} and \textit{Expectancy-Value Theory}, which capture intrinsic versus extrinsic drivers of engagement~\citep{yue2022international,gladstone2022situated,loh2019we}.  
Together, these constructs define how agents interpret classroom signals, prioritize goals, and sustain effort over time, even under identical task conditions.

Each student agent is initialized with a structured personality profile encoded in JSON format, including its trait vector, cognitive preferences, and motivational disposition. During execution, these parameters are accessed primarily in the \textit{Plan} stage of the PCA loop, guiding intention formulation and prompt conditioning. For instance, a highly agreeable agent tends to produce affirming or cooperative responses during group discussion, whereas a low-conscientiousness agent is more likely to show task-avoidant tendencies. Reflective learners adopt deliberate monitoring strategies and gradual adjustment, while impulsive learners often exhibit frequent shifts and inconsistent participation.

By embedding these individualized profiles into the cognitive control cycle, CIE generates agents that combine \textbf{behavioral realism} with \textbf{systematic variability}. The heterogeneity among agents promotes the simulation of authentic classroom dynamics, such as uneven participation or role differentiation. Moreover, these embedded traits provide interpretable signals for teacher agents, enabling differentiated instruction and context-sensitive feedback strategies. In this way, the personality-conditioned cognition module supports both fine-grained behavioral modeling and pedagogical analysis, bridging psychological theory with computational simulation.

\begin{tcolorbox}[breakable, colback=gray!5!white, colframe=gray!60!black, title=Virtual Student Profile: \WangFang, fonttitle=\bfseries]

\vspace{0.5em}
\textbf{Basic Information:} 
\begin{itemize}
  \item Age: 13 \quad Gender: Female
  \item Spawn Position: (5, 1)
  \item Perception Config: Range = 3, Max Tracked Items = 5
\end{itemize}

\vspace{0.5em}
\textbf{Learning Engagement:} 
\begin{itemize}
    \item Behavior = Head-down reading
    \item Emotion = Positive
    \item Cognition = Understanding
\end{itemize}

\vspace{0.5em}
\textbf{Personality Traits (Big Five):}
\begin{itemize}
  \item Personality Type: High Neuroticism
  \item Neuroticism: 0.9, Extraversion: 0.5, Agreeableness: 0.5, Openness: 0.5, Conscientiousness: 0.5
  
\end{itemize}

\vspace{0.5em}
\textbf{Cognitive Style:}
\begin{itemize}
  \item Field Independent
  \item Reflective
\end{itemize}

\vspace{0.5em}
\textbf{Learning Motivation:} High

\vspace{0.5em}
\textbf{Class Role:} Academic-Oriented Student

\vspace{0.5em}
\textbf{Behavior Profile:}
\begin{itemize}
  \item \textbf{Language Style:} Hesitant, emotionally influenced; uses fillers like “um”, backtracks responses, shows fragmented expression.
  \item \textbf{Class Behavior:} Anxious under pressure; avoids raising hand but listens attentively in groups; completes work slowly but with logical clarity and strong motivation.
  \item \textbf{Learning Preference:} Prefers independent tasks, reflective practice; likes using notes or diagrams for organizing knowledge; enjoys self-paced deep exploration.
  \item \textbf{Teacher Guidance:} Reduce performance anxiety via timely feedback; provide structured task breakdowns and staged learning goals; encourage written or non-verbal responses.
  \item \textbf{Role Description:} Acts as knowledge summarizer in groups; organizes ideas clearly; prefers private channels for expressing ideas to build confidence.
\end{itemize}

\end{tcolorbox}

\subsubsection{Personality-driven Style Modulation}

To ensure that agents maintain trait-aligned consistency in language and behavior generation across multi-turn interactions, \textbf{CIE} introduces a dual-module mechanism: the \textbf{Style Modulator} and the \textbf{Cognitive Generator}. This design balances the generalization strength of large language models (LLMs) with the need for personality-conditioned expression, thereby ensuring that each virtual student consistently exhibits distinctive communicative styles and cognitive tendencies during simulation.

\textbf{Style Modulator.}  
We fine-tuned four mainstream Chinese LLMs—InternVL~\citep{chen2024internvl}, Qwen~\citep{bai2023qwen}, MiniCPM~\citep{hu2024minicpm}, and LLaVa~\citep{li2024llava}—on more than 6,000 utterances collected from real-world junior secondary Chinese classroom interactions. These data span five instructional phases (e.g., lesson introduction, explanation) and two core questioning modes. Among the candidates, InternVL achieved the highest scenario-grounded performance and was selected as the primary generator. Fine-tuning was performed using the parameter-efficient LoRA method~\citep{hu2021lora}. To enhance persona-aware generation, we further introduced a tri-label annotation format—capturing personality traits, behavioral tendencies, and task contexts—allowing the model to learn discriminative linguistic patterns associated with specific learner archetypes.

\textbf{Cognitive Generator.}  
At each stage of the metacognitive loop—\textit{Plan, Monitor, Regulate}—the system issues role-specific prompts that drive context-aware decision-making. These prompts integrate the agent’s personality profile, current instructional signals, and historical behavioral traces, thereby enabling dynamic adjustment of cognitive strategies. Prompt templates are separately designed for student and teacher agents, supporting the complete perception–cognition–action loop described in earlier sections.

Overall, this dual-module design not only enhances the realism and stability of personality-conditioned expression but also provides an interpretable framework for modeling diverse educational behaviors in simulated classrooms.

\begin{tcolorbox}[breakable,enhanced,colback=blue!5!white, colframe=blue!50!black,
  title=Plan Prompt for Virtual Student Agent, fonttitle=\bfseries,fontupper=\normalsize]

You are a student named \textit{\{self.name\}}, and your character profile is \textit{(\{self.profile\})}. Based on your persona and the following contextual information, reflect on your learning goal and behavioral plan in the current class session:

\begin{itemize}
    \item Current physical distance: \textit{\{self.profile[dist\_teacher]\}, \{self.profile[dist\_blackboard]\}}
  \item Teacher’s current behavior: \textit{\{teacher\_behavior\}}
  \item Teacher’s emotional tone: \textit{\{teacher\_emotion\}}
  \item Teacher’s instructional content: \textit{\{teacher\_content\}}
  \item Other students’ previous standing responses: \textit{\{stu\_response\}}
  \item Previous student questions to the teacher: \textit{\{stu\_request\}}
  \item Learning status of neighboring classmates: \textit{\{neighbors\}}
  \item Relevant instructional objects: \textit{\{objects\}}
  \item Your recent learning experiences and regulation suggestions: \textit{\{memory[:3]\}}
  \item Your personality type: \textit{\{self.personality\_type\}}
  \item Your learning behavior style: \textit{\{self.class\_behavior\}}
  \item Your classroom role: \textit{\{self.class\_role\}}
  \item Current lesson content: \textit{\{self.shared\_state['lesson\_content']\}}
\end{itemize}

Based on the above information and your personality traits, describe your behavioral plan using a first-person reasoning chain. The behavior must fall into one of the following four categories:

\begin{itemize}
  \item \textbf{Active behaviors:} taking notes, raising hand
  \item \textbf{Passive behaviors:} listening attentively, reading silently, reading aloud
  \item \textbf{Interactive behaviors:} side talk with peers, asking the teacher questions
  \item \textbf{Disengaged behaviors:} sleeping on desk, chatting with others
\end{itemize}

\textbf{Instructions:}
\begin{enumerate}
  \item Carefully consider the teacher's state, peer learning status, your persona, and recent self-regulation experience.
  \item Your behavior should align with your habitual classroom pattern—e.g., if you are disengaged by nature, plan accordingly, unless your self-regulation history indicates change.
  \item Output must follow a first-person reasoning chain—concise and limited to a single sentence.
  \item Select \textbf{only one final behavior} from the list (e.g., taking notes / raising hand / listening attentively / reading silently / reading aloud / side talk / asking questions / sleeping / chatting).
  \item Strictly follow the output format below:
  
  The teacher is asking about spring-related poetry, with a calm tone. My neighbors are actively raising their hands XXXX. \\
  Final behavior: side talk with peers
\end{enumerate}

\end{tcolorbox}

\begin{tcolorbox}[breakable,enhanced,colback=blue!5!white, colframe=blue!50!black,
  title=Monitor Prompt for Virtual Student Agent, fonttitle=\bfseries,fontupper=\normalsize]

You are a student named \textit{\{self.name\}}, and your character profile is \textit{(\{self.profile\})}. Based on your persona and the following contextual information, reflect on the current instructional situation and evaluate whether your understanding and behavior are appropriate:

\begin{itemize}
  \item Teacher’s behavior: \textit{\{teacher\_behavior\}}
  \item Neighboring student states: \textit{\{neighbors\}}
  \item Your previous plan: \textit{\{plan\_output\}}
  \item Recent memory fragments: \textit{\{memory\}}
  \item Your current behavior: \textit{\{action\_result.get('behavior', 'unknown')\}}
  \item Feedback received: \textit{\{action\_result.get('response', 'none')\}}
  \item Your personality type: \textit{\{self.personality\_type\}}
  \item Your classroom behavior habit: \textit{\{self.class\_behavior\}}
\end{itemize}

Based on the information above, assess your current emotional and cognitive state under the behavior you just performed. Emotions should be classified into three categories: \textbf{positive}, \textbf{negative}, or \textbf{confused}.  
Cognitive states should follow Bloom's taxonomy and be selected from: \textbf{Remembering}, \textbf{Understanding}, \textbf{Applying}, \textbf{Analyzing}, \textbf{Evaluating}, \textbf{Creating}.

\textbf{Instructions:}
\begin{enumerate}
  \item Briefly monitor your behavioral process using a first-person perspective.
  \item Output your emotional and cognitive category results strictly in the following format:
  
  I think XXX \\
  Emotion: XXX \\
  Cognition: XXX
\end{enumerate}

\end{tcolorbox}

\begin{tcolorbox}[breakable,enhanced,colback=blue!5!white, colframe=blue!50!black,
  title=Regulate Prompt for Virtual Student Agent, fonttitle=\bfseries,fontupper=\normalsize]

You are a student named \textit{\{self.name\}}, and your character profile is \textit{\{self.profile\}}. Based on your persona, your recent monitoring reflection \textit{(\{monitor\_output\})}, memory state \textit{(\{memory\})}, and the teacher’s feedback \textit{(\{teacher\_feedback\})}, reflect on what adjustment strategy you should adopt next (e.g., asking a question, taking notes, communicating with peers, etc.).

\textbf{Additional contextual information:}
\begin{itemize}
  \item Teacher’s current behavior: \textit{\{teacher\_behavior\}}
  \item Your personality type: \textit{\{self.personality\_type\}}
  \item Your classroom learning habits: \textit{\{self.class\_behavior\}}
  \item Current social interaction partners: \textit{\{self.shared\_state["social\_interaction"]\}}
  \item Peers who rejected your conversation attempts: \textit{\{social\_request\_reject\}}
\end{itemize}

\textbf{Instructions:}
\begin{enumerate}
  \item Briefly analyze how you would like to adjust your learning strategy in the next step.
  \item The output should be concise and stated in the first person, using only one sentence.
\end{enumerate}

\textbf{Example:} "I got distracted just now, and the teacher called on me; I hope to refocus and pay better attention."

\end{tcolorbox}

\begin{tcolorbox}[breakable,enhanced,colback=blue!5!white, colframe=blue!50!black,
  title=Prompt for Virtual Teacher: Reconstructing Lesson Plan into Instructional Phases, fonttitle=\bfseries,fontupper=\normalsize]

You are an experienced and professional middle school Chinese language teacher. Below is the instructional plan for the lesson titled \textit{\{self.lesson\_id\}}. Please read the lesson plan carefully and reconstruct it according to the required structure.

\textbf{Task:}  
Carefully review the content of the \textit{\{self.lesson\_id\}} lesson plan. Reorganize the instructional content into five standard teaching phases:  
\textbf{Lesson Introduction, New Content Instruction, Knowledge Consolidation, In-Class Practice, and Lesson Summary}.

The lesson consists of a total of 30 time steps. For each phase:
\begin{itemize}
  \item Provide a concise summary of the instructional content to be covered, written in paragraph form (no bullet points), within 50 characters (or equivalent).
  \item Allocate a specific number of time steps to each phase.
\end{itemize}

\textbf{Lesson Plan Content:}
\begin{verbatim}
--- Lesson Plan Start ---
{self.lesson_plan_text}
--- Lesson Plan End ---
\end{verbatim}

\textbf{Please strictly follow the output format below:}

\begin{verbatim}
Lesson Introduction: XXXX, Steps: XXX
New Content Instruction: XXXX, Steps: XXX
Knowledge Consolidation: XXXX, Steps: XXX
In-Class Practice: XXXX, Steps: XXX
Lesson Summary: XXXX, Steps: XXX
\end{verbatim}

\end{tcolorbox}

\begin{tcolorbox}[breakable,enhanced,colback=blue!5!white, colframe=blue!50!black,
  title=Prompt for Virtual Teacher: Instructional Step Planning, fonttitle=\bfseries,fontupper=\normalsize]

You are an experienced and professional middle school Chinese language teacher. You are currently teaching the course \textit{\{self.lesson\_id\}} and are now in the instructional phase titled \textit{"\{teaching\_phase\}"}. Please plan the instructional content for this specific phase.

This phase is expected to span \textit{\{total\_steps\}} instructional steps.
Below is a brief summary of the content you are expected to teach in this phase: \textit{\{teaching\_phase\}}

\textbf{Instructions:}
\begin{enumerate}
  \item Break down the above instructional content into \textbf{\{total\_steps\}} individual teaching steps.
  \item Each step should consist of one concise sentence (no more than 20 Chinese characters or equivalent length in English).
  \item Ensure that each step aligns clearly with the goal of the current instructional phase.
  \item Avoid repetition, vague statements, or logical leaps between steps.
  \item Use the following format for output:

  \begin{verbatim}
  1. xxx
  2. xxx
  3. xxx
  ...
  \end{verbatim}
\end{enumerate}

\end{tcolorbox}

\begin{tcolorbox}[breakable,enhanced,colback=blue!5!white, colframe=blue!50!black,
  title=Prompt for Virtual Teacher: Selecting Instructional Behavior Type, fonttitle=\bfseries,fontupper=\normalsize]

You are an experienced and professional middle school Chinese language teacher. You are currently teaching the lesson \textit{\{self.lesson\_id\}}, and are now in the instructional phase \textit{"\{phase\}"}. The content planned for the current time step is: \textit{\{teaching\_step\_content\}}.

Please determine the most appropriate instructional behavior type based on the following contextual information:

\begin{itemize}
  \item Student participation, emotional, and cognitive states: \textit{\{perception\_result['teaching']['statistics']\}}
  \item Your most recent teaching feedback: \textit{\{teacher\_feedback\}}
  \item Student self-regulation output: \textit{\{student\_regulation\}}
  \item Your last behavior category: \textit{\{category\}}
\end{itemize}

\textbf{Instructions:}
\begin{enumerate}
  \item Consider whether to maintain or shift from your previous behavior category (\textit{\{category\}}).
  \item It is generally preferred to vary your instructional strategy across consecutive time steps for richer pedagogical dynamics.
  \item Select one behavior category from the following list:
  \begin{itemize}
    \item Classroom Instruction (e.g., lecturing, giving directions)
    \item Classroom Interaction (e.g., expressing emotion, praise, incorporating student input, asking questions, giving criticism, organizing discussion)
    \item Classroom Behavior Management (e.g., addressing students sleeping or chatting)
  \end{itemize}
  \item \textbf{Important:} Output only the final selected category name without numbers.  
  \textit{For example:} \texttt{Classroom Interaction}
\end{enumerate}

\end{tcolorbox}

\begin{tcolorbox}[breakable,enhanced,colback=blue!5!white, colframe=blue!50!black,
  title=Prompt for Virtual Teacher: Selecting Fine-Grained Teaching Action, fonttitle=\bfseries,fontupper=\normalsize]

You are an experienced and professional middle school Chinese language teacher. You are currently teaching the lesson \textit{\{self.lesson\_id\}} and are in the instructional phase \textit{"\{phase\}"}. The planned instructional content for the current time step is: \textit{\{teaching\_step\_content\}}.

You have decided to perform a \textit{\{category\}} type of teaching behavior. Based on the following contextual information, please select one fine-grained instructional action that aligns with your selected behavior type.

\begin{itemize}
  \item Current phase: \textit{\{phase\}}
  \item Behavior type selected: \textit{\{category\}}
  \item Student participation and emotional state: \textit{\{perception\_result['teaching']['statistics']\}}
\end{itemize}

\textbf{Available fine-grained behaviors by category:}
\begin{itemize}
  \item \textbf{Classroom Instruction:} lecturing, giving directions
  \item \textbf{Classroom Interaction:} expressing emotion, praising, adopting student input, asking questions, giving criticism, organizing group discussion
  \item \textbf{Classroom Behavior Management:} addressing students sleeping, addressing students chatting
\end{itemize}

\textbf{Instructions:}
\begin{itemize}
  \item Select only one fine-grained behavior that best suits the context.
  \item Do not include any numbering or extra explanation.
  \item \textit{Example:} lecturing
\end{itemize}

\end{tcolorbox}

\begin{tcolorbox}[breakable,enhanced,colback=blue!5!white, colframe=blue!50!black,
  title=Prompt for Virtual Teacher: Generating Instructional Utterance, fonttitle=\bfseries,fontupper=\normalsize]

You are an experienced and professional middle school Chinese language teacher. You are currently teaching the lesson \textit{\{self.lesson\_id\}}, and are in the instructional phase \textit{"\{phase\}"}. The instructional content planned for this time step is approximately: \textit{\{teaching\_step\_content\}}.  
Your selected teaching behavior is: \textit{\{behavior\}}. Please generate the instructional utterance you will deliver to students based on the following context:

\begin{itemize}
  \item Current instructional phase: \textit{\{phase\}}
  \item Teaching behavior: \textit{\{behavior\}}
  \item Planned teaching content for this time step: \textit{\{teaching\_step\_content\}}
  \item Current student states: \textit{\{perception\_result['teaching']['statistics']\}}
  \item Recent content already covered: \textit{\{[h['content'] for h in history[-3:]]\}}
  \item Text material being taught: \textit{\{self.shared\_state['lesson\_content']\}}
\end{itemize}

\textbf{Instructions:}
\begin{enumerate}
  \item Your utterance must logically follow previously delivered content and align with the current teaching goal.
  \item \textbf{Avoid repeating prior statements.}
  \item If your behavior is \textbf{lecturing}, you may deliver up to 5 informative sentences focused on knowledge delivery.
  \item For all other behavior types, limit the output to \textbf{2–3 sentences}.
  \item You may refer to specific students based on what you know about them (e.g., call them by name), except when the behavior is \textbf{organizing classroom discussion}. In that case, pose an open-ended prompt to all students, optionally setting up a collaborative or competitive task.
\end{enumerate}

\textbf{Please output only the generated utterance (no metadata).}

\end{tcolorbox}

\begin{tcolorbox}[breakable,enhanced,colback=blue!5!white, colframe=blue!50!black,
  title=Prompt for Virtual Teacher: Selecting Instructional Emotion, fonttitle=\bfseries,fontupper=\normalsize]

You are an experienced and professional middle school Chinese language teacher. You are currently teaching the lesson \textit{\{self.lesson\_id\}} and are in the instructional phase \textit{"\{phase\}"}. The teaching content for the current time step is: \textit{\{content\}}, and your current instructional behavior is: \textit{\{behavior\}}.

Please determine the most appropriate emotional tone for this moment based on the following student emotional state:

\begin{itemize}
  \item Student emotional distribution: \textit{\{emo\}}
\end{itemize}

\textbf{Instructions:}
\begin{itemize}
  \item Choose one emotional tone from the following three options:
  \begin{itemize}
    \item \textbf{Encouraging}
    \item \textbf{Critical}
    \item \textbf{Neutral}
  \end{itemize}
  \item Please output \textbf{only one} of the three tones, with no additional explanation.
  \item \textit{Example output: Encouraging}
\end{itemize}

\end{tcolorbox}

To ensure stylistic and semantic consistency across multi-turn dialogues, \textbf{CIE} introduces a mechanism that aligns dialogue history with continuity of style. Specifically, the output generated at each \textit{Regulate} stage is propagated forward into the subsequent \textit{Plan} stage, thereby preserving coherence in both personality expression and cognitive trajectory. Response generation follows a hybrid pipeline: a LoRA-tuned LLM first produces a personality-aligned draft, which is then refined by GPT-4 to guarantee pedagogical validity and cognitive plausibility. This dual-stage strategy integrates personality intent, cognitive structure, and contextual awareness into a unified and adaptive response process.

\begin{tcolorbox}[breakable,enhanced,colback=blue!5!white, colframe=blue!50!black,
  title=Prompt for Personality-Driven Style Modulation, fonttitle=\bfseries,fontupper=\normalsize]

You are an assistant that can revise student responses based on their personality characteristics while preserving their individual speaking style.

The student’s personality type is: \textit{\{self.personality\_type\}}. Below is a one-sentence sample response in their characteristic style. Your task is to refine it based on the student's persona without altering the style.

\textbf{Student Information:}
\begin{itemize}
  \item Name: \textit{\{self.name\}}
  \item Personality Type: \textit{\{self.personality\_type\}}
  \item Language Style: \textit{\{self.language\_style\}}
  \item Classroom Behavior: \textit{\{self.class\_behavior\}}
\end{itemize}

\textbf{Context:}
\begin{itemize}
  \item Teacher's question: \textit{\{query\}}
  \item The student plans to raise their hand to respond.
  \item Their drafted response: \textit{\{draft\}}
  \item Learning plan for this time step: \textit{\{plan\}}
\end{itemize}

\textbf{Instructions:}
\begin{enumerate}
  \item Evaluate whether the drafted response is reasonable. If it is not, you may disregard it and instead generate a new answer based on the student’s personality traits.
  \item Middle school students typically speak concisely—your revised response should follow similar length and tone as the sample.
  \item Your final output should:
  \begin{itemize}
    \item Maintain the student's original speaking style;
    \item Reflect their personality and classroom behavior;
    \item Align with the current instructional context.
  \end{itemize}
  \item \textbf{Output only the revised response.}
\end{enumerate}

\end{tcolorbox}

\subsection{Detailed Description of Social Situatedness}
\label{subapp:cie-i}

\subsubsection{Theoretical Expansion of IRF Paradigm}

The Initiation–Response–Feedback (IRF) model, first proposed by Sinclair and Coulthard, remains a cornerstone in classroom discourse analysis~\citep{waring2009moving,rustandi2017analysis}. It organizes interaction into three stages: the teacher \textit{initiates} with prompts or questions (I), the student \textit{responds} verbally or behaviorally (R), and the teacher provides \textit{feedback} (F) in the form of evaluation or elaboration. While concise and widely applicable, this formulation is limited in capturing the cognitive and social dynamics of modern classrooms.

Under constructivist and inquiry-oriented pedagogies, students act as reflective learners rather than passive recipients~\citep{walker2015understanding,renninger2024exploring}. They monitor, regulate, and socially negotiate their learning in response to both internal states and external cues—capacities insufficiently represented in the original IRF framework.

To address this gap, \textbf{CIE} extends IRF into a four-phase structure, termed IRF-R, by introducing a \textbf{Regulation} stage. After receiving feedback, students engage in metacognitive processing: reassessing their performance, adjusting goals, and modifying strategies in light of emotional state, peer interaction, and task relevance. In this expanded cycle, \textit{Initiation} stimulates attention and motivation, \textit{Response} generates verbal or behavioral engagement, \textit{Feedback} reinforces or redirects cognition, and \textit{Regulation} transforms feedback into adaptive behavior.

The IRF-R paradigm thus supports multi-turn interaction loops that conceptualize learning as a continuous cycle of stimulation, expression, feedback, and self-adjustment. It enables teacher agents to track not only immediate responses but also downstream learning adjustments, thereby improving the interpretability of student behavior and supporting deeper trajectories of engagement.

\subsubsection{Implementation of Social Prioritization}

In the \textbf{CIE} multi-agent system, student agents must simultaneously handle instructional prompts from teacher agents and spontaneous peer-initiated interactions during lessons. To resolve conflicts among these competing inputs, we introduce a \textbf{social prioritization mechanism} inspired by gated decision control. This ensures that at each time step, every agent responds to the interaction with the highest \textit{pedagogical relevance} (see Alg.~\ref{alg:social-gating}).

\begin{algorithm}[H]
\caption{\textbf{Social Priority Gating Mechanism}}
\label{alg:social-gating}
\KwIn{Local perception $s_{i_p}^t$ of agent $i$, shared state $S$, social threshold $\theta$}
\KwOut{Behavioral decision $a_i^t$}
\BlankLine
\uIf{$\textbf{TeacherRequestExists}(i, s_{i_p}^t)$}{
    $a_i^t \leftarrow$ respond to teacher (e.g., ``stand and answer'')\;
    \textbf{RejectSocialRequest}$(i, \text{`Teacher Priority'})$\;
}
\uElseIf{\textbf{SocialRequestExists}$(i, S)$}{
    $r_i^t \leftarrow \textbf{GetSocialRequest}(i, S)$\;
    $H_{i,j}, S_{i,j} \leftarrow \textbf{AnalyzeChatHistory}(i, r_i^t.\text{from\_id})$\;
    $W_i^t \leftarrow \textbf{ComputeIntention}(P_i, C_i, R_i, H_{i,j}, S_{i,j})$\;
    \uIf{$W_i^t \geq \theta$}{
        $a_i^t \leftarrow r_i^t.\text{type}$\;
        $r_i^t.\text{status} \leftarrow \text{accepted}$\;
    }
    \Else{
        $a_i^t \leftarrow \text{Self-Initiated Learning}$\;
        \textbf{RejectSocialRequest}$(i, \text{`Low Intention'})$\;
    }
}
\Else{
    $a_i^t \leftarrow \text{Self-Initiated Learning}$\;
}
\Return $a_i^t$;
\end{algorithm}

Interaction priorities are structured into three tiers.  
First, \textbf{teacher requests} (e.g., direct questioning or task assignments) override all other interactions, placing the student agent in an uninterruptible execution state.  
Second, if no teacher request is present, the agent evaluates \textbf{peer requests} and \textbf{group discussions} (e.g., side chats, peer questions). Here, an LLM-based reasoning process integrates personality traits, task context, and interaction history to compute a \textit{social willingness score} $W_i^t \in [0,1]$. If $W_i^t \geq \theta$ (e.g., $\theta=0.6$), the request is accepted; otherwise, it is rejected and the agent returns to self-regulated learning.  
Third, in the absence of external input, the agent continues self-initiated planning and behavior execution.

To ensure transparent tracking, all requests are logged in the shared interaction state pool with one of three tags: \textbf{Pending} (awaiting response), \textbf{Accepted} (engagement initiated; dialogue content logged for future reasoning), or \textbf{Rejected} (declined without side effects; agent resumes autonomous learning). This tagging protocol enables fine-grained regulation of peer dialogue without disrupting the core instructional flow.

Overall, the gating-based prioritization mechanism preserves instructional coherence while still allowing socially grounded behaviors to emerge adaptively and contextually in classroom simulations.

\subsubsection{Social Intention Function}

In multi-agent classroom environments, modeling whether a student agent is willing to accept peer-initiated interactions is crucial for simulating realistic social behavior. To this end, \textbf{CIE} introduces a language model–driven \textbf{Social Intention Function}, which dynamically determines willingness based on the current instructional context. This function integrates five factors—personality traits, learning confidence, task relevance, historical interaction frequency, and social closeness—into a context-aware decision process.

To ensure educational interpretability, each factor is anchored in established psychological theories:  
\begin{itemize}
    \item \textbf{Personality Match ($P$)}: Following the \textit{Big Five Personality Theory}, students high in extraversion tend to be socially responsive, while those high in neuroticism are more likely to avoid interaction~\citep{zhao2006big,john1999big}.
    \item \textbf{Current Learning Confidence ($C$)}: Based on \textit{Bandura’s Self-Efficacy Theory}, confidence in task performance directly shapes one’s propensity for social engagement~\citep{schunk2016self,bandura1977analysis}.
    \item \textbf{Task Relevance ($R$)}: Grounded in \textit{Situated Cognition} and \textit{Constructivist Learning Theory}, this factor assesses whether a social request aligns with the ongoing instructional objective~\citep{wilson2000situated,hedegaard1998situated}.
    \item \textbf{Historical Interaction Frequency ($H$)} and \textbf{Social Closeness ($S$)}: Informed by \textit{Social Identity Theory}, these capture group belonging and accumulated positive peer interactions~\citep{hogg2016social,ellemers2012social}.
\end{itemize}

Each component is scored within the range $[0,100]$ using GPT-4 via fine-tuned prompts, represented as $P, C, R, H, S$. Simultaneously, the system generates a context-dependent weight vector:
\begin{equation}
\boldsymbol{\alpha} = [\alpha_1, \alpha_2, \alpha_3, \alpha_4, \alpha_5], \quad \text{subject to} \quad \sum_{i=1}^{5} \alpha_i = 1.
\end{equation}

The model also outputs a short justification for each weight to enhance interpretability. The final \textbf{Social Intention Score} is computed as:
\begin{equation}
W = \alpha_1 P + \alpha_2 C + \alpha_3 R + \alpha_4 H + \alpha_5 S.
\end{equation}

If $W \geq 0.6$ (default threshold), the student accepts the request and enters a dialogue; otherwise, the request is rejected and the agent resumes self-directed learning.

This mechanism functions not merely as a scoring model, but as a cognitively and socially grounded reflection of student behavior. It captures a learner’s “social rationality” and “regulatory capacity” across tasks and roles, enhancing both the behavioral realism and interpretability of CIE’s agent-based interaction model.

\subsection{Detailed Description of Temporal Dynamics}
\label{subapp:cie-e}

\subsubsection{Theoretical Foundations of Agent State Progression and Regulation}

In \textbf{CIE}, the learning process of virtual students is conceptualized not as isolated one-step reactions but as a temporally extended trajectory unfolding across multiple time steps. This trajectory reflects progressive \textbf{cognitive development}, \textbf{emotional regulation}, and \textbf{behavioral adaptation}. To capture these dynamics, we introduce a dual mechanism of \textbf{State Progression} and \textbf{State Regulation}, grounded in established theories of educational psychology, thereby modeling agent-level development across lessons and instructional phases.

\textbf{Cognitive development} is anchored in Bloom’s Taxonomy~\citep{krathwohl2002revision,forehand2010bloom} and Bruner’s Spiral Curriculum~\citep{clark2010jerome,joseph2021bruner}. Bloom’s hierarchy delineates a progression from lower- to higher-order cognition:
\[
\text{Remembering} \;\rightarrow\; \text{Understanding} \;\rightarrow\; \text{Applying} \;\rightarrow\; \text{Analyzing} \;\rightarrow\; \text{Evaluating} \;\rightarrow\; \text{Creating}.
\]
In CIE, each student agent’s cognitive output is annotated accordingly at each step, enabling temporal trend analysis. Complementarily, Bruner’s spiral principle emphasizes cyclical revisiting of concepts with increasing complexity, allowing longitudinal tracking of knowledge deepening, reinforcement, and occasional regression.

\textbf{Emotional and behavioral adaptation} is informed by Emotion Regulation Theory~\citep{gross2008emotion,gross1999emotion}, Self-Determination Theory (SDT)~\citep{deci2017self,deci2012self}, and Expectancy–Value Theory~\citep{wigfield1994expectancy,wigfield2000expectancy}. Gross highlights that learners regulate emotions through strategies such as support-seeking, withdrawal, or task switching. In CIE, these processes shape the \texttt{Regulate} module, where agents evaluate affective states to guide strategic adjustment. SDT introduces autonomy, competence, and relatedness as motivational parameters; these are embedded into CIE’s motivation and social intention functions, influencing persistence and behavioral shifts across time. Expectancy–Value perspectives further explain how perceived value and anticipated success jointly determine sustained engagement.

\textbf{Teacher–student feedback dynamics} build upon Hattie and Timperley’s Feedback Model~\citep{hattie2007power} and Vygotsky’s Sociocultural Theory~\citep{lantolf2000introducing,scott2013sociocultural}. The former stresses that effective feedback triggers metacognitive reassessment and regulation beyond error correction; in CIE, such feedback modifies both goal-setting and emotional states in subsequent \texttt{Plan} phases. Vygotsky’s theory adds a social dimension: knowledge construction is mediated by interaction, and teacher or peer feedback indirectly shapes confidence, regulation, and discourse strategies.

In summary, the integration of state progression and regulation mechanisms allows CIE to simulate learning as a temporally grounded, theory-consistent developmental process. Virtual students are thus modeled not as reactive output devices but as evolving educational agents whose longitudinal behavioral trajectories provide interpretable evidence of learning dynamics and cognitive pathway development.

\subsubsection{Teacher-controlled Instructional Pacing}

\setcounter{AlgoLine}{0}
\begin{algorithm}[H]
\caption{Teacher-controlled Instructional Phase Pacing}
\label{alg:teacher-control}
\KwIn {Current instructional phase $\text{Phase}_k$, student state set $\{s_i^t\}_{i=1}^n$, step count $t$}
\KwOut {Next instructional phase $\text{Phase}_{k+1}$}
Initialize total steps per phase: $T \leftarrow 30$ \;

Initialize current step index: $t \leftarrow 1$ \;

Initialize teacher policy: $\pi_T$ \;

\While{$t \leq T$}{
    Observe group student state $S^t = \{s_i^t\}_{i=1}^n$ \;
    
    Execute teacher action $a_T^t \sim \pi_T(S^t, \text{Phase}_k)$ \;
    
    Broadcast action $a_T^t$ to all student agents \;
    
    Collect responses and update $S^{t+1}$ \;
    
    $t \leftarrow t + 1$ \;
}

Evaluate completion rate: $r_k \leftarrow \text{Evaluate}(S^t)$ \;

Determine transition: $\text{Phase}_{k+1} \leftarrow \text{Transition}(\text{Phase}_k, S^t, r_k)$ \;

\Return $\text{Phase}_{k+1}$

\end{algorithm}

In multi-step instructional simulations, effective pacing control is essential not only for synchronizing with students’ learning rhythms but also for maintaining interactional coherence and managing cognitive load. To address this challenge, the \textbf{CIE} framework incorporates a \textbf{Teacher Agent Rhythm Control Module}, which dynamically governs the progression of instructional phases throughout a lesson.

Each lesson is preconfigured into five canonical instructional phases (e.g., introduction, explanation, consolidation), further decomposed into discrete teaching steps. During simulation, the teacher agent adaptively decides whether to continue, delay, or advance phase transitions based on real-time classroom signals rather than following a fixed timeline.

This decision process operates through a \textbf{Perception–Cognition–Action (PCA)} loop. At each step $t$, the teacher agent first perceives aggregated student states $S^t = \{s_i^t\}_{i=1}^n$ (e.g., participation density, emotional distribution, cognitive indicators). In the cognition stage, it evaluates whether the current phase $\text{Phase}_k$ should be sustained or transitioned by integrating lesson plan constraints, system logs, and recent feedback. Finally, the action stage executes the pacing decision, broadcasting teacher actions $a_T^t$ and updating the global state.

The pacing controller thus implements a data-driven mechanism that ensures phase boundaries remain pedagogically aligned and interpretable across the session. Each time step is explicitly logged with its transition rationale, enabling post-hoc analysis and iterative refinement of instructional design. The full scheduling logic is summarized in Alg.~\ref{alg:teacher-control}.

\subsection{Memory Mechanisms for Agent Cognition}
\label{subapp:cie-m}

\subsubsection{Theoretical Foundations of Memory Mechanisms}

In cognitive psychology and the learning sciences, memory mechanisms are central to understanding how learners encode, retain, and retrieve knowledge for decision-making~\citep{terry2017learning,sprenger1999learning}. To simulate this process in virtual agents, CIE implements a dual-layer memory architecture inspired by the classical \textit{Working Memory–Long-Term Memory} model. This design ensures both real-time responsiveness and cross-session continuity in agent cognition and behavior.

\textbf{Working memory} provides short-term storage of salient instructional information during ongoing sessions—such as teacher actions, peer interactions, emotional states, and cognitive indicators. These records are stored in a global $\texttt{shared\_state}$ structure, which is updated at each time step $t$. This memory layer enables the real-time execution of the \textit{Perception–Cognition–Action (PCA)} loop, featuring high temporal resolution and frequent access, thereby forming the basis for moment-to-moment decision-making.

\textbf{Long-term memory}, in parallel, functions as a persistent repository of knowledge accumulation and behavioral trajectories. Implemented as a structured database, it logs each student’s historical records across lessons, including cognitive progression, emotional trends, and task engagement. At the beginning of each new session, the long-term memory is reloaded into the $\texttt{shared\_state}$, enabling agents to adapt based on prior experiences. This mechanism supports \textit{experience-informed learning} and retrospective reasoning across multiple episodes.

By integrating these two layers, CIE models both the ephemeral and cumulative aspects of learning. The separation between rapidly evolving working states and persistent knowledge encoding ensures that agents can respond fluidly to the immediate instructional context while also continuously adapting to their developmental history. The following sections provide a detailed description of the design and flow between these two memory layers.

\subsubsection{Implementation of Short-term Shared Memory}

In CIE, the short-term memory mechanism—termed \textbf{short-term shared memory}—defines a unified interaction state pool, $\texttt{shared\_state}$, which facilitates high-frequency, real-time information exchange among all agents, including both students and teachers. Drawing inspiration from the psychological construct of working memory, this module temporarily stores task-relevant perceptual information and cognitive-affective states within the current instructional phase, thereby enabling synchronized decision-making across agents.

The $\texttt{shared\_state}$ consists of several structured components, initialized at the start of each session:

\begin{itemize}
    \item \textbf{Teacher state}: records the teacher’s current behavior, utterances, and emotional tone.
    \item \textbf{Request pools (instructional \& social)}: manage the lifecycle of agent-to-agent interaction requests, including initiation, acceptance, and rejection.
    \item \textbf{Student logs}: capture each agent’s cognitive level, emotional state, and behavior trace at every time step $t$.
    \item \textbf{Spatial state}: represents agent-specific surroundings and perceptible objects, grounding interactions in physical context.
    \item \textbf{Task \& group context}: contains the active lesson plan, group composition, and instructional content segments.
    \item \textbf{Interaction history}: maintains continuity of dialogue rounds and the evolution of collaborative strategies.
    \item \textbf{Long-term memory reference}: integrates episodic summaries from previous lessons to initialize experience-informed behaviors.
\end{itemize}

This state pool is updated dynamically at each instructional step. For example, when a student initiates a peer request, the entry is appended to the social pool; when the teacher provides feedback, the teacher state is refreshed; when students respond, their logs are updated with behavioral and cognitive annotations. This continuous update cycle ensures that all agents maintain a synchronized representation of the evolving classroom environment.

In summary, the short-term shared memory functions as the temporal backbone of coordinated multi-agent interaction in CIE. By supporting real-time perception and regulation within the Perception–Cognition–Action loop, it enables adaptive and coherent decision-making in high-frequency educational scenarios.

\subsubsection{Implementation of Long-term Memory}

To sustain behavioral continuity and cumulative cognitive development across sessions, CIE incorporates a structured \textbf{long-term memory system} consisting of two SQLite databases: \texttt{student\_memory.db} and \texttt{teacher\_memory.db}. These databases respectively record student learning trajectories and teacher instructional behaviors over time.

\textbf{Student Memory}: The \texttt{student\_memory.db} contains a \texttt{long\_term\_memory} table with the following schema:

\begin{itemize}
    \item \texttt{student\_id}: unique identifier for each student agent.
    \item \texttt{event\_type}: type of record (e.g., \textit{Cognitive Planning}, \textit{Monitoring}, \textit{Regulation}, \textit{Behavioral Record}).
    \item \texttt{event\_content}: natural language logs produced during metacognitive stages, including goal setting or strategic reflection.
    \item \texttt{timestamp}: temporal marker that enables reconstruction of student-specific learning sequences.
\end{itemize}

This structure ensures traceability of self-regulated learning activities, forming a temporally grounded chain along the \textit{Perception–Cognition–Action} loop.

\textbf{Teacher Memory}: The \texttt{teacher\_memory.db} mirrors this design, maintaining a \texttt{long\_term\_memory} table that logs instructional records at the phase level:

\begin{itemize}
    \item \texttt{teacher\_id}: unique identifier for the teacher agent.
    \item \texttt{lesson\_id, phase}: identifiers of the lesson and instructional phase.
    \item \texttt{event\_type}: type of teacher behavior (e.g., \textit{Instructional Planning}, \textit{Feedback}).
    \item \texttt{content}: natural language descriptions of teacher intentions, evaluations, and scaffolding strategies.
    \item \texttt{timestamp}: time of execution, supporting longitudinal modeling of instructional dynamics.
\end{itemize}

A particular focus is placed on feedback chains (e.g., student question $\rightarrow$ teacher evaluation $\rightarrow$ student regulation), which provide the basis for context-aware instructional planning in future sessions and adaptive modeling of scaffolding behaviors.

Together, these two databases enable bidirectional memory transfer: teacher agents analyze longitudinal patterns in student performance, while student agents draw upon accumulated knowledge, goals, and feedback to guide future behavior. The long-term memory system thus allows CIE to simulate instructional interaction as an evolving and authentic temporal process.

\subsubsection{Memory Interaction Flow}

To support cognitive development and behavioral adaptation across instructional sessions, CIE implements a structured \textbf{Memory Interaction Flow} that coordinates short-term and long-term memory. This mechanism enables student agents to accumulate, apply, and evolve learning experiences within and across sessions, thereby sustaining coherence and continuity in personalized learning trajectories.

At the start of each session, the system loads individual long-term memory summaries—such as prior performance, cognitive tendencies, and emotional traits—from the database into the \texttt{shared\_state}. These values immediately inform in-session planning and response generation.

During instruction, student agents continuously update the short-term memory at each time step by recording their actions, peer interactions, and teacher feedback. These records ensure real-time context awareness and guide micro-level cognitive regulation.

At the end of a session, the system aggregates time-step records into structured learning summaries. These include updated cognitive markers, emotional trajectories, and selected behavior patterns, which are written back into the long-term memory for use in subsequent sessions.

This continuous memory flow enables student agents to engage in feedback-informed, data-driven self-regulation. Over time, they develop individualized learning patterns reflective of authentic developmental trajectories in classroom environments. The complete execution logic of this flow is outlined in Alg.~\ref{alg:memory-flow}.

\setcounter{AlgoLine}{0}
\begin{algorithm}[H]
\caption{\textbf{Memory Interaction Flow for Agent $i$}}
\label{alg:memory-flow}
\KwIn{Agent ID $i$, shared state $S$, current step $t$, total steps $T$}
\KwOut{Updated shared state $S$}
\BlankLine

\uIf{$t = 1$}{
    $L_i \leftarrow \text{DATABASE.retrieve\_long\_term\_summary}(i)$\;
    
    $S[\text{longterm\_summary}][i] \leftarrow L_i$\;
}
\BlankLine

$S[\text{interaction\_log}][i][t] \leftarrow$ \{

\quad $\text{teacher\_interactions}:~s_i^t.\text{teacher\_content},$ \\
\quad $\text{social\_interactions}:~s_i^t.\text{social\_requests},$ \\
\quad $\text{student\_responses}:~s_i^t.\text{stu\_response},$ \\
\quad $\text{student\_requests}:~s_i^t.\text{stu\_request},$ \\
\quad $\text{environment\_context}:~s_i^t.\text{visible\_items},$ \\
\quad $\text{emotional\_states}:~s_i^t.\text{teacher\_emotion},$ \\
\quad $\text{cognitive\_states}:~\text{MEMORY.retrieve\_cog}()$ 

\}

\BlankLine
\uIf{$t = T$}{

    $M_i \leftarrow \text{Summarize\_Session}(S[\text{interaction\_log}][i])$\;
    
    \text{DATABASE.store\_long\_term\_summary}$(i, M_i)$\;

    $S[\text{longterm\_summary}][i].\text{pre\_lesson\_summary} \leftarrow M_i.\text{summary}$\;
    
    $S[\text{longterm\_summary}][i].\text{pre\_lesson\_portrait} \leftarrow M_i.\text{portrait}$\;
    
    $S[\text{longterm\_summary}][i].\text{pre\_lesson\_regulation} \leftarrow M_i.\text{regulation}$\;
    
    $S[\text{longterm\_summary}][i].\text{teacher\_evaluation} \leftarrow M_i.\text{teacher\_eval}$\;

}

\Return $S$\;

\end{algorithm}

\clearpage

\section{Detailed Information for Experimental Design and Evaluation}
\label{sec:detailed-exp}

\subsection{Experiment Setup and Configuration}

To investigate how virtual students respond cognitively, emotionally, and socially across diverse instructional contexts, we developed a Chinese language classroom simulation integrating genre diversity, phased pedagogy, and structured interaction protocols.  

For materials, three representative texts from the junior secondary curriculum were selected: the lyrical prose \textit{Spring} (Zhu Ziqing), the foreign fable \textit{The Emperor’s New Clothes}, and the argumentative essay \textit{Dedication and Joy}. These texts differ in linguistic style, cognitive demand, and emotional resonance, supporting heterogeneous tasks such as expressive description, character analysis with moral reasoning, and logical argumentation for critical debate.  

Interaction was structured through an extended IRF model. By adding a regulation phase, we formed an ``I--R--F--Regulate'' loop: the teacher initiates, students respond, the teacher provides feedback, and students regulate through reflection or social actions (e.g., questioning, discussion, or strategy adjustment). This design preserves traditional instructional dialogue while enhancing agents’ behavioral and emotional expressiveness.  

Each lesson comprised five pedagogical phases---introduction, instruction, consolidation, practice, and summarization---mapped to approximately 30 steps but dynamically adjusted by the teacher’s policy to about 36 steps per session, depending on task completion and engagement signals.  

The agent architecture followed a two-tier design: GPT-4 was responsible for natural language generation and reasoning, while a fine-tuned InternVL model modulated style. Each student was encoded as $[p_i; c_i; m_i]$---personality, cognitive style, and motivation---and combined with phase, memory, and context in prompt templates. Prompts were configured with temperature $=0.5$, max tokens $=512$, top-$p=0.9$, and frequency penalty $=0.2$.  

Experiments were run on a server with H20-NVLink GPUs (96GB VRAM) and 200GB RAM. Each inference step averaged 25 seconds, and a full class of six students plus one teacher lasted 1--2 hours. All session data---including cognitive, emotional, and behavioral annotations---were stored in a MongoDB backend for longitudinal continuity analysis.  

To simulate learner diversity, six virtual student archetypes were designed based on the Big Five personality model, motivational theory, and cognitive style literature. These archetypes reflect typical learner profiles in real classrooms and enable the evaluation of interactional variance and pedagogical robustness.

\tcbset{
  colback=gray!0!white,
  colframe=black!15,
  width=\columnwidth,
  sharp corners,
  boxrule=0.5pt,
  fonttitle=\bfseries,
  before skip=5pt,
  after skip=5pt,
}

\begin{tcolorbox}[breakable,title = \WangFang]
\textbf{Age:} 13 \\
\textbf{Gender:} Female \\
\textbf{Personality:} High Neuroticism \\
\textbf{Class Role:} Academic Student \\
\textbf{Learning Motivation:} High \\
\textbf{Class Behavior:} Often hesitant but gradually contributes well-structured summaries and personal reflections during class. \\
\textbf{Cognitive Style:} Field-dependent \\
\textbf{Thinking Tendency:} Reflective
\end{tcolorbox}

\begin{tcolorbox}[breakable,title = \ZhangJie]
\textbf{Age:} 14 \\
\textbf{Gender:} Male \\
\textbf{Personality:} High Extraversion \\
\textbf{Class Role:} Academic Student \\
\textbf{Learning Motivation:} High \\
\textbf{Class Behavior:} Actively initiates discussions, shares opinions confidently, and frequently stands up to respond or ask questions. \\
\textbf{Cognitive Style:} Field-independent \\
\textbf{Thinking Tendency:} Reflective
\end{tcolorbox}

\begin{tcolorbox}[breakable,title = \ZhangYan]
\textbf{Age:} 13 \\
\textbf{Gender:} Female \\
\textbf{Personality:} High Agreeableness \\
\textbf{Class Role:} Academic Student \\
\textbf{Learning Motivation:} High \\
\textbf{Class Behavior:} Frequently engages in peer interaction, supports others' ideas, and shows strong cooperative communication. \\
\textbf{Cognitive Style:} Field-independent \\
\textbf{Thinking Tendency:} Reflective
\end{tcolorbox}

\begin{tcolorbox}[breakable,title = \LiWei]
\textbf{Age:} 14 \\
\textbf{Gender:} Male \\
\textbf{Personality:} Low Openness \\
\textbf{Class Role:} Discussion Student \\
\textbf{Learning Motivation:} High \\
\textbf{Class Behavior:} Leads group discussion with structured logic, seeks consensus, and promotes balanced participation. \\
\textbf{Cognitive Style:} Field-dependent \\
\textbf{Thinking Tendency:} impulsive
\end{tcolorbox}

\begin{tcolorbox}[breakable,title = \LiuLi]
\textbf{Age:} 13 \\
\textbf{Gender:} Female \\
\textbf{Personality:} Low Openness \\
\textbf{Class Role:} Off-task Student \\
\textbf{Learning Motivation:} Low \\
\textbf{Class Behavior:} Easily distracted in class, often avoids eye contact, but occasionally responds with emotional expressions. \\
\textbf{Cognitive Style:} Field-dependent \\
\textbf{Thinking Tendency:} Impulsive
\end{tcolorbox}

\begin{tcolorbox}[breakable,title = \ZhangTao]
\textbf{Age:} 14 \\
\textbf{Gender:} Male \\
\textbf{Personality:} Low Conscientiousness \\
\textbf{Class Role:} Off-task Student \\
\textbf{Learning Motivation:} Low \\
\textbf{Class Behavior:} Tends to disengage from class tasks, shows low participation, and often chats about irrelevant topics. \\
\textbf{Cognitive Style:} Field-dependent \\
\textbf{Thinking Tendency:} Impulsive
\end{tcolorbox}

\clearpage
\subsection{Evaluation Metric}
\label{sub:evaluation-matrix}
\begin{itemize}
    \item \textbf{IRF rate (Initiation–Response–Feedback rate).} 
    Defined as the proportion of time steps in which a complete IRF cycle is realized, i.e., a teacher initiation ($I_T^t$), followed by a student response ($R_S^t$), and completed with teacher feedback ($F_T^t$). This metric captures instructional alignment between teacher and students.

    \item \textbf{Behavior categories (ICAP framework).} 
    We classify student behaviors into four categories: 
    \emph{Active} (e.g., answering questions, note-taking), 
    \emph{Passive} (e.g., listening, reading), 
    \emph{Interactive} (e.g., peer discussion, asking questions), and 
    \emph{Off-Class} (e.g., off-task, inattentive).

    \item \textbf{Cognition categories (Bloom’s taxonomy).} 
    Cognitive engagement is divided into two levels: 
    \emph{Lower-order cognition} (Remember, Understand) and 
    \emph{Higher-order cognition} (Apply, Analyze, Evaluate, Create).

    \item \textbf{Emotion categories.} 
    Emotional states are coded into three classes: 
    \emph{Positive}, 
    \emph{Negative}, and 
    \emph{Confused}.
\end{itemize}

\clearpage

\subsection{Experiment I}


\begin{table}[h!]
  \centering
  \begin{threeparttable}
  \caption{\textbf{IRF distribution for three text genres across four environments.} 
  Values are relative frequencies of Initiation (I), Response (R), and Feedback (F). 
  \textbf{IRF\_rate} denotes the overall completion ratio.}
  \label{tab:full-IRF}
  \setlength{\tabcolsep}{4pt}
  \small
  \begin{tabular}{p{2.6cm}lcccccc}
    \toprule
    \textbf{Text Genre} & \textbf{Env.} & \textbf{Steps} & \textbf{I} & \textbf{R} & \textbf{F} & \textbf{IRF\_rate} \\
    \midrule
    \multirow{4}{*}{\textit{Lyrical Prose}} 
      & Lecture     & 37 & 0.514 & 0.167 & 0.275 & 0.432 \\
      & Collab      & 41 & 0.439 & 0.179 & 0.321 & 0.293 \\
      & Round       & 39 & 0.410 & 0.154 & 0.282 & 0.282 \\
      & Real Class  & 37 & 0.513 & --    & 0.703 & 0.486 \\
    \midrule
    \multirow{4}{*}{\textit{Argumentative Essay}}
      & Lecture     & 36 & 0.556 & 0.194 & 0.282 & 0.639 \\
      & Collab      & 40 & 0.375 & 0.213 & 0.367 & 0.475 \\
      & Round       & 31 & 0.516 & 0.215 & 0.355 & 0.548 \\
      & Real Class  & 36 & 0.417 & --    & 0.583 & 0.417 \\
    \midrule
    \multirow{4}{*}{\textit{Foreign Fiction}}
      & Lecture     & 33 & 0.364 & 0.253 & 0.475 & 0.455 \\
      & Collab      & 33 & 0.242 & 0.247 & 0.394 & 0.303 \\
      & Round       & 37 & 0.324 & 0.189 & 0.351 & 0.378 \\
      & Real Class  & 33 & 0.367 & --    & 0.515 & 0.367 \\
    \bottomrule
  \end{tabular}
  \begin{tablenotes}
    \footnotesize
    \item IRF = Initiation, Response, and Feedback ratio in dialogue.
    \item Real Class = Real classroom environment.
  \end{tablenotes}
  \end{threeparttable}
\end{table}


\begin{table}[h!]
  \centering
  \caption{\textbf{Distribution of students’ behavior, cognition, and emotion across different text genres and environments.}}
  \label{tab:beh-cog-emo}
  \setlength{\tabcolsep}{2pt}
  \small
  \begin{tabular}{p{1.2cm}p{2.7cm}cccccccccc}
    \toprule
    \textbf{Env.} & \textbf{Text} & \textbf{B\_Aac} & \textbf{B\_Pas} & \textbf{B\_Int} & \textbf{B\_Off} & \textbf{E\_Pos} & \textbf{E\_Con} & \textbf{E\_Neg} & \textbf{C\_Low} & \textbf{C\_High} \\
    \midrule
    \multirow{3}{*}{Lecture} 
      & \textit{Lyrical Prose}      & 0.157 & 0.222 & 0.398 & 0.222 & 0.509 & 0.463 & 0.028 & 0.875 & 0.125 \\
      & \textit{Argumentative Essay}& 0.200 & 0.276 & 0.324 & 0.200 & 0.481 & 0.519 & 0.000 & 0.790 & 0.210 \\
      & \textit{Foreign Fiction}    & 0.161 & 0.260 & 0.396 & 0.182 & 0.510 & 0.484 & 0.005 & 0.792 & 0.208 \\
    \midrule
    \multirow{3}{*}{Collab} 
      & \textit{Lyrical Prose}      & 0.171 & 0.204 & 0.454 & 0.171 & 0.483 & 0.517 & 0.000 & 0.942 & 0.058 \\
      & \textit{Argumentative Essay}& 0.303 & 0.205 & 0.316 & 0.175 & 0.547 & 0.312 & 0.141 & 0.697 & 0.303 \\
      & \textit{Foreign Fiction}    & 0.188 & 0.203 & 0.313 & 0.297 & 0.609 & 0.359 & 0.031 & 0.578 & 0.422 \\
    \midrule
    \multirow{3}{*}{Round} 
      & \textit{Lyrical Prose}      & 0.154 & 0.154 & 0.461 & 0.232 & 0.482 & 0.518 & 0.000 & 0.807 & 0.193 \\
      & \textit{Argumentative Essay}& 0.189 & 0.200 & 0.383 & 0.228 & 0.511 & 0.406 & 0.083 & 0.794 & 0.206 \\
      & \textit{Foreign Fiction}    & 0.204 & 0.269 & 0.227 & 0.301 & 0.616 & 0.296 & 0.088 & 0.861 & 0.139 \\
    \bottomrule
  \end{tabular}
\end{table}


\begin{figure}[h!]
    \centering
    \includegraphics[width=\textwidth]{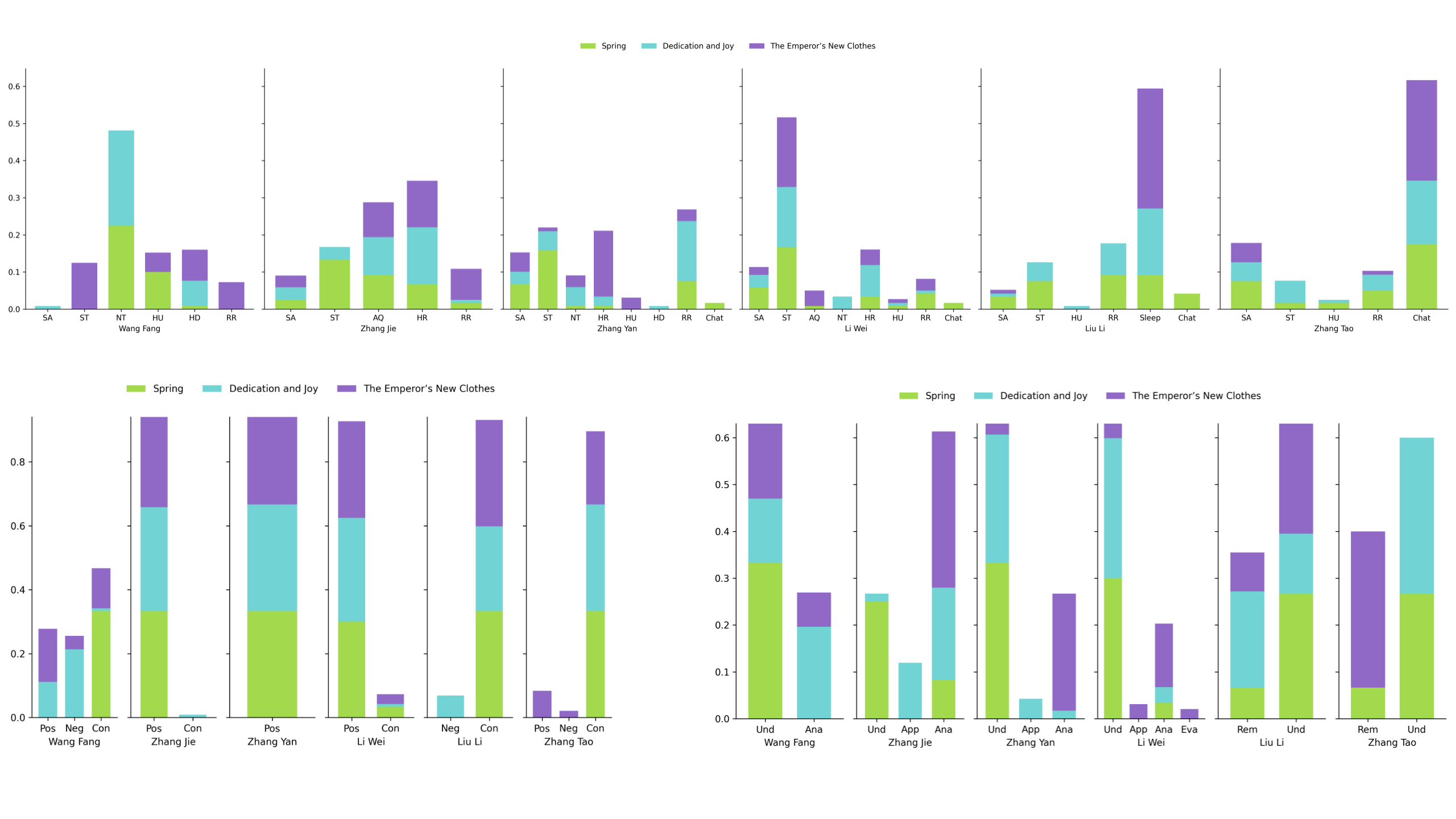}
    \vspace{-15pt}
    \caption{Student behavior, emotion and cognition frequencies across three lessons in collab environment.}
    \label{fig:individual_collab}
\end{figure}

\begin{figure}[h!]
    \centering
    \includegraphics[width=\textwidth]{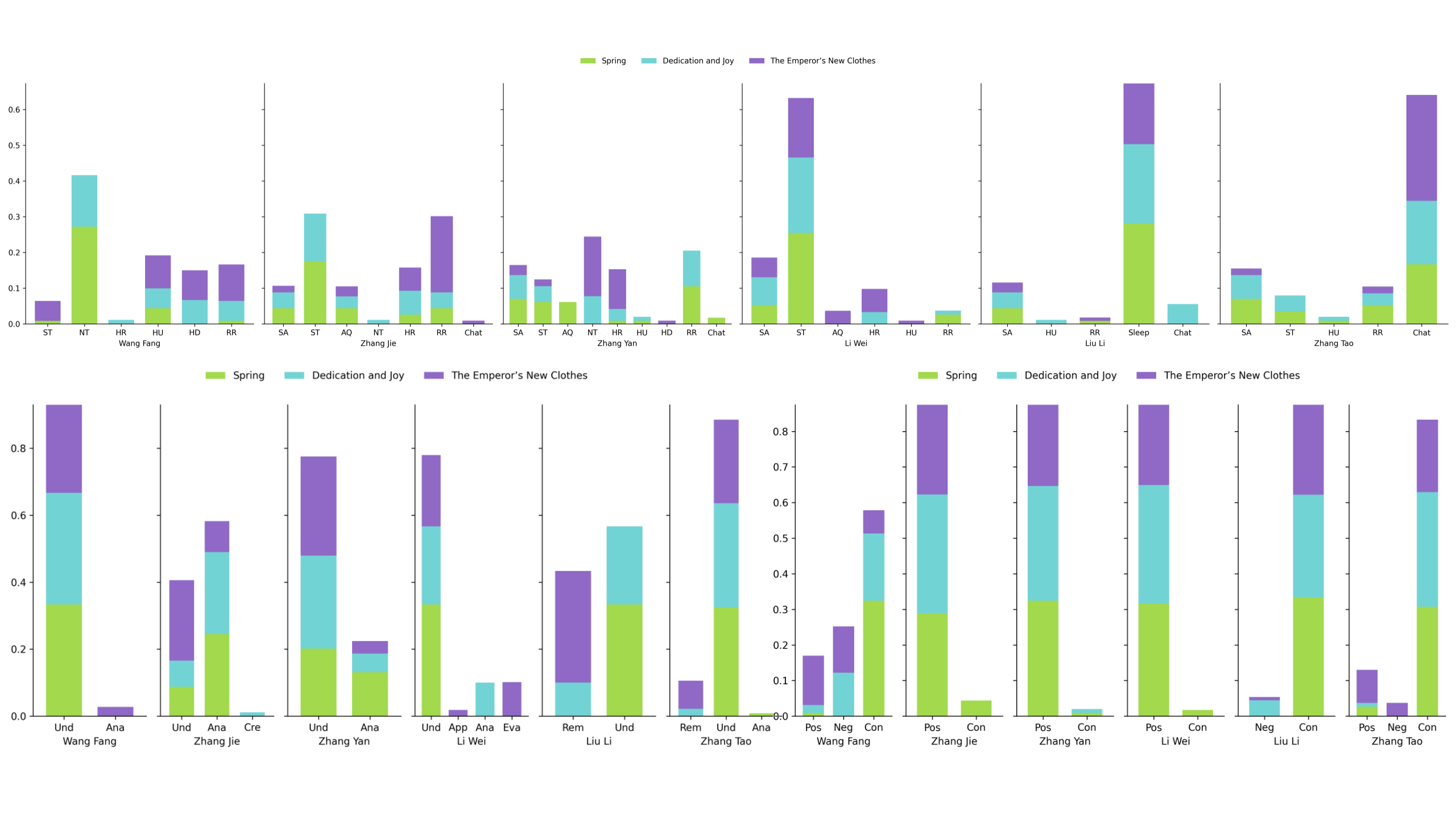}
    \vspace{-15pt}
    \caption{Student behavior, emotion and cognition frequencies across three lessons in round environment.}
    \label{fig:individual_round}
\end{figure}


\begin{table}[h!]
  \centering
  \caption{\textbf{Ablation experiments across different frameworks.} 
  Results are reported on three text genres. 
  ``nolocal'' denotes removing local interaction rules, while ``nocog'' denotes removing cognitive mechanisms.}
  \label{tab:ablation}
  \setlength{\tabcolsep}{2pt}
  \small
  \begin{tabular}{p{2.7cm}p{1.7cm}cccccccccc}
    \toprule
    \textbf{Text Genre} & \textbf{Framework} & \textbf{B\_Aac} & \textbf{B\_Pas} & \textbf{B\_Int} & \textbf{B\_Off} & \textbf{E\_Pos} & \textbf{E\_Con} & \textbf{E\_Neg} & \textbf{C\_Low} & \textbf{C\_High} \\
    \midrule
    \multirow{3}{*}{\textit{Lyrical Prose}} 
      & Ours     & 0.157 & 0.222 & 0.398 & 0.222 & 0.509 & 0.463 & 0.028 & 0.875 & 0.125 \\
      & nolocal  & 0.321 & 0.195 & 0.224 & 0.260 & 0.557 & 0.301 & 0.142 & 0.890 & 0.110 \\
      & nocog    & 0.315 & 0.212 & 0.216 & 0.257 & 0.505 & 0.351 & 0.144 & 0.631 & 0.369 \\
    \midrule
    \multirow{3}{*}{\textit{Argumentative Essay}} 
      & Ours     & 0.200 & 0.276 & 0.324 & 0.200 & 0.481 & 0.519 & 0.000 & 0.790 & 0.210 \\
      & nolocal  & 0.221 & 0.225 & 0.329 & 0.225 & 0.523 & 0.302 & 0.176 & 0.883 & 0.117 \\
      & nocog    & 0.229 & 0.233 & 0.233 & 0.304 & 0.563 & 0.429 & 0.008 & 0.479 & 0.521 \\
    \midrule
    \multirow{3}{*}{\textit{Foreign Fiction}} 
      & Ours     & 0.161 & 0.260 & 0.396 & 0.182 & 0.510 & 0.484 & 0.005 & 0.792 & 0.208 \\
      & nolocal  & 0.210 & 0.271 & 0.295 & 0.224 & 0.524 & 0.390 & 0.086 & 0.748 & 0.252 \\
      & nocog    & 0.266 & 0.306 & 0.158 & 0.270 & 0.509 & 0.468 & 0.023 & 0.176 & 0.824 \\
    \bottomrule
  \end{tabular}
\end{table}

To complement the main findings of Experiment I, this section provides additional analyses using IRF statistics, behavior–cognition–emotion (BCE) distributions, individual-level visualizations, and ablation studies.

First, the full \textbf{IRF statistics across genres and environments} (Tab.~\ref{tab:full-IRF}) show that simulated classrooms maintain comparable IRF completion rates with real classes. For instance, in lyrical prose, Lecture/Collab/Round yield IRF\_rates between 0.28 and 0.43, versus 0.486 in the real classroom. Argumentative essays and foreign fiction also exhibit distributions of similar magnitude. These findings substantiate the main-text claim that simulated discourse structures align closely with real classrooms, while minor deviations (e.g., higher rates in Lecture for argumentative essays) reflect the annotation protocol.

Second, the \textbf{BCE distribution} (Tab.~\ref{tab:beh-cog-emo}) further illustrates classroom ecology. Overall, lower-order cognition dominates (e.g., Lecture–LP $C\_Low=0.875$), positive and confused emotions prevail, and negative affect remains low. Environment effects vary by genre: in foreign fiction, Collab produces higher-order cognition ($C\_High=0.422$), while Round leads to more off-task behavior ($B\_Off=0.301$) and less interaction ($B\_Int=0.227$). In contrast, Lecture settings in lyrical prose and argumentative essays show more passive participation and lower-order cognition. These results highlight a genre–environment interaction that systematically shapes classroom dynamics.

Third, the \textbf{individual-level visualizations} (Fig.~\ref{fig:individual_collab} and Fig.~\ref{fig:individual_round}) confirm personality-driven stability. Highly extraverted or conscientious students sustain active engagement and positive affect, whereas low-openness or low-conscientiousness students display more off-task behavior, low-level cognition, and frequent confusion. This consistency across lessons aligns with the main-text analysis.

Finally, the \textbf{ablation experiments} (Tab.~\ref{tab:ablation}) disentangle the role of key modules. Removing localized interaction rules (\texttt{nolocal}) increases negative emotions and undermines affective realism. Removing cognitive mechanisms (\texttt{nocog}) exaggerates higher-order reasoning and overly active behaviors, deviating from gradual, student-like learning. Together, these results confirm that style modulation preserves emotional plausibility, while the cognitive layer ensures educational consistency.

In sum, this section corroborates three central claims of Experiment I: the \textbf{authenticity of discourse structures}, the \textbf{stability of personality-driven patterns}, and the \textbf{necessity of modular design}, while also emphasizing that genre–environment interactions play a critical role in shaping simulated classroom dynamics.

\clearpage

\subsection{Experiment II}


\begin{figure}[h!]
    \centering
    \includegraphics[width=0.5\textwidth]{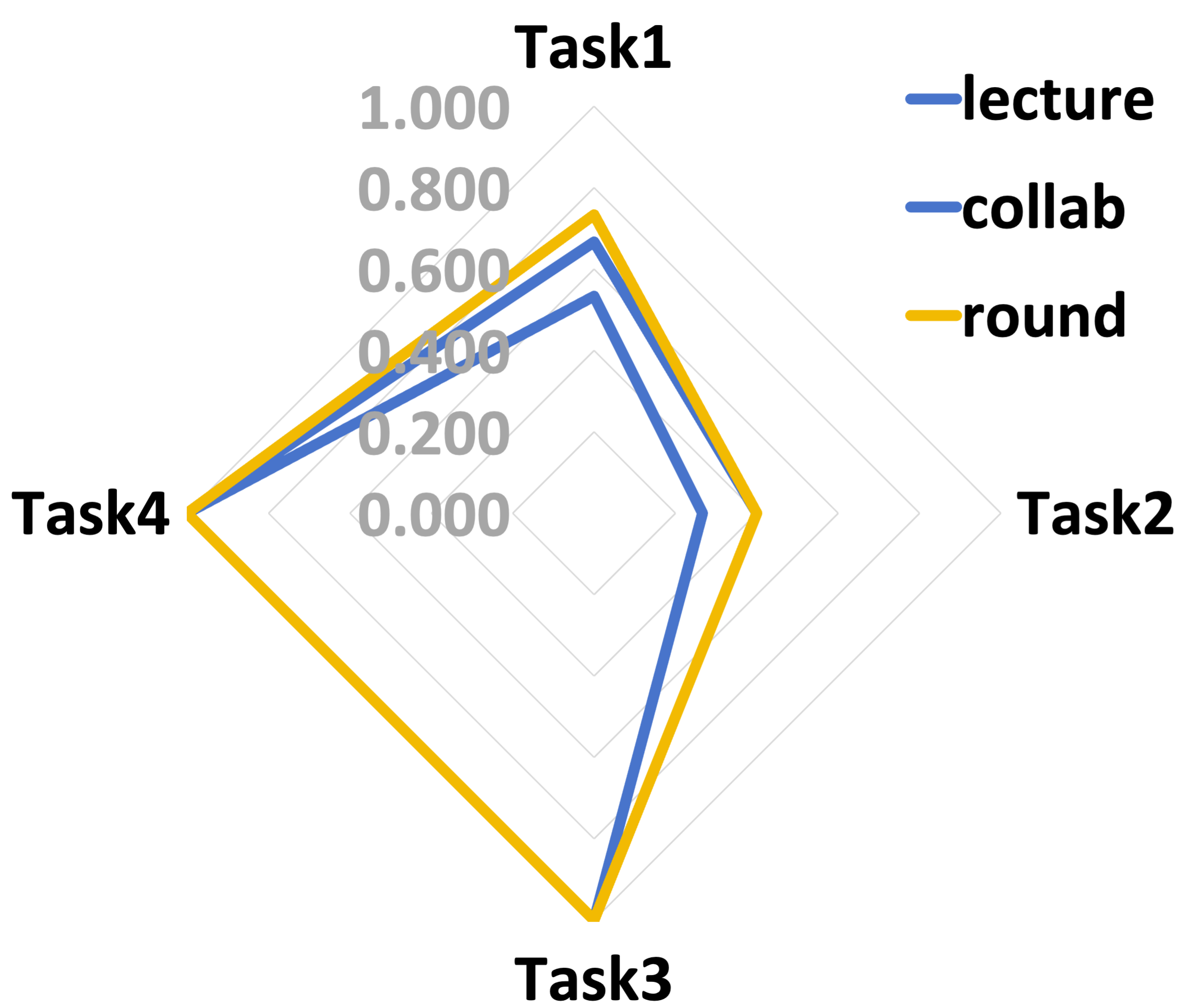}
    \caption{\textbf{Human–Agent Interaction Across Four Tasks.} Results show strong alignment with personality traits and robust instructional control, confirming that EduVerse enables seamless integration of human participants while preserving realistic classroom dynamics.}
\label{fig:rader}
\end{figure}

\begin{table}[htbp]
\centering
\caption{\textbf{Distribution of students’ network centrality indicators in classroom interaction contexts.} 
Values are normalized to $[0,1]$; \textbf{Degree} = in+out centrality.}
\label{tab:centrality_full}
\setlength{\tabcolsep}{4pt}
\small
\begin{tabular}{p{3cm}p{1.5cm}lcccc}
\toprule
\textbf{Text} & \textbf{Env.} & \textbf{Student} & \textbf{In} & \textbf{Out} & \textbf{Degree} & \textbf{Betweenness} \\
\midrule
\multirow{17}{*}{\textit{Foreign Fiction}} 
  & \textit{Lecture} & \WangFang & 0.20 & 0.20 & 0.40 & 0    \\
  &                  & \ZhangTao & 0.20 & 0.40 & 0.60 & 0.20 \\
  &                  & \LiWei    & 0.20 & 0.20 & 0.40 & 0    \\
  &                  & \LiuLi    & 0.40 & 0.40 & 0.80 & 0.30 \\
  &                  & \ZhangYan & 0.40 & 0.40 & 0.80 & 0.40 \\
  &                  & \ZhangJie & 0.40 & 0.20 & 0.60 & 0.30 \\
  \cmidrule{2-7}
  & \textit{Collab}  & \WangFang & 0.25 & 0.25 & 0.50 & 0     \\
  &                  & \LiWei    & 0.25 & 0.25 & 0.50 & 0     \\
  &                  & \ZhangTao & 0.25 & 0.50 & 0.75 & 0.083 \\
  &                  & \ZhangJie & 0.25 & 0.00 & 0.25 & 0     \\
  &                  & \ZhangYan & 0.25 & 0.25 & 0.50 & 0     \\
  \cmidrule{2-7}
  & \textit{Round}   & \WangFang & 0.40 & 0.20 & 0.60 & 0.15 \\
  &                  & \LiWei    & 0.20 & 0.20 & 0.40 & 0    \\
  &                  & \ZhangTao & 0.40 & 0.40 & 0.80 & 0.25 \\
  &                  & \ZhangJie & 0.20 & 0.20 & 0.40 & 0    \\
  &                  & \ZhangYan & 0.00 & 0.40 & 0.40 & 0    \\
  &                  & \LiuLi    & 0.20 & 0.00 & 0.20 & 0    \\
\midrule
\multirow{17}{*}{\textit{Lyrical Prose}} 
  & \textit{Lecture} & \LiWei    & 0.20 & 0.20 & 0.40 & 0    \\
  &                  & \ZhangJie & 0.40 & 0.40 & 0.80 & 0.15 \\
  &                  & \LiuLi    & 0.20 & 0.20 & 0.40 & 0    \\
  &                  & \ZhangYan & 0.20 & 0.20 & 0.40 & 0    \\
  &                  & \ZhangTao & 0.20 & 0.40 & 0.60 & 0.10 \\
  &                  & \WangFang & 0.20 & 0.00 & 0.20 & 0    \\
  \cmidrule{2-7}
  & \textit{Collab}  & \LiWei    & 0.25 & 0.25 & 0.50 & 0     \\
  &                  & \LiuLi    & 0.25 & 0.25 & 0.50 & 0     \\
  &                  & \ZhangTao & 0.25 & 0.25 & 0.50 & 0     \\
  &                  & \ZhangYan & 0.50 & 0.50 & 1.00 & 0.167 \\
  &                  & \ZhangJie & 0.25 & 0.25 & 0.50 & 0     \\
  \cmidrule{2-7}
  & \textit{Round}   & \WangFang & 0.20 & 0.20 & 0.40 & 0.10 \\
  &                  & \LiuLi    & 0.20 & 0.00 & 0.20 & 0    \\
  &                  & \LiWei    & 0.20 & 0.40 & 0.60 & 0.10 \\
  &                  & \ZhangJie & 0.20 & 0.40 & 0.60 & 0.10 \\
  &                  & \ZhangTao & 0.20 & 0.20 & 0.40 & 0    \\
  &                  & \ZhangYan & 0.40 & 0.20 & 0.60 & 0.10 \\
\bottomrule
\end{tabular}
\end{table}

\begin{table}[htbp]
\centering
\caption{\textbf{Human–agent interaction success rates across four tasks and three environments.} 
Values represent completion ratios (0–1).}
\label{tab:human-agent}
\setlength{\tabcolsep}{6pt}
\small
\begin{tabular}{lccc}
\toprule
\textbf{Task} & \textbf{Lecture} & \textbf{Collab} & \textbf{Round} \\
\midrule
\textit{Task 1: Peer chatting}             & 0.533 & 0.667 & 0.733 \\
\textit{Task 2: Peer academic response}    & 0.267 & 0.400 & 0.400 \\
\textit{Task 3: Teacher answering}         & 1.000 & 1.000 & 1.000 \\
\textit{Task 4: Teacher intervention}      & 1.000 & 1.000 & 1.000 \\
\bottomrule
\end{tabular}
\end{table}

In the supplementary analyses of Experiment II, we further validated the stability and realism of EduVerse through network centrality indicators and human–agent interaction tasks. As shown in Tab.~\ref{tab:centrality_full}, students displayed distinct role patterns across environments. For the \textit{Foreign Fiction} lesson, in the Lecture setting \ZhangYan (Degree=0.80, Betweenness=0.40) and \LiuLi (Degree=0.80, Betweenness=0.30) emerged as core participants, while \ZhangTao exhibited stronger initiative through higher outward connections (Out=0.40). 
In the Collab setting, interactions became more reciprocal, with most students maintaining a Degree of around 0.50, but \ZhangTao rose to 0.75 with a nonzero betweenness (0.083), serving as a connector. In the Round setting, the network further decentralized, with bridging roles distributed: \ZhangTao (Degree=0.80, Betweenness=0.25) and \WangFang (Betweenness=0.15) acted as key connectors, while \ZhangYan showed a distinctive one-directional output pattern (In=0.00, Out=0.40). 
For the \textit{Lyrical Prose} lesson, \ZhangJie (Degree=0.80, Betweenness=0.15) was central in Lecture, \ZhangYan dominated in  Collab (Degree=1.00, Betweenness=0.167), and Round produced a more balanced structure with several students sharing moderate centralities (Degree=0.60, Betweenness=0.10). 
These results indicate that while classroom environment reshapes group structures, personality-driven individual traits remain relatively stable and interpretable.

Human–agent interaction tests, reported in Tab.~\ref{tab:human-agent}, further demonstrate EduVerse’s adaptability. In Task 1 (peer chatting), the socially inclined \ZhangTao responded in most cases, with success rates of 0.533, 0.667, and 0.733 in Lecture, Collab, and Round respectively, and occasionally initiated chats himself in collaborative or roundtable settings. By contrast, in Task 2 (peer academic response), the conscientious \ZhangJie rarely engaged during class, yielding lower success rates (0.267–0.400). In Tasks 3 (teacher answering) and 4 (teacher intervention), success rates consistently reached 1.000 across all environments, confirming that the teacher agent reliably answered questions and actively intervened in off-task behaviors.

Taken together, these supplementary results confirm that EduVerse not only reproduces realistic group dynamics and individual differences but also sustains authentic role-driven behavior when human users are integrated. Socially oriented students show high willingness to engage, academically conscientious students remain task-focused, and teacher agents reliably maintain instructional order. This underscores the robustness and applicability of EduVerse in complex interaction and human-in-the-loop classroom scenarios.

\clearpage

\subsection{Experiment III}

Each student’s time series of state annotations is processed to compute pairwise transitions between consecutive time steps. A transition is labeled as \textbf{positive} if the subsequent state reflects a higher level of engagement, affect, or cognition based on the predefined state hierarchy (e.g., behavior: \texttt{disengaged} = 0, ..., \texttt{interactive} = 3).
Formally, let $s_t$ be the state at time $t$ with scalar priority $P(s_t) \in \mathbb{Z}^{\geq 0}$. A positive transition occurs when:

\begin{equation}
P(s_{t+1}) > P(s_t)
\end{equation}

Define:

\begin{equation}
T = \sum_{t=1}^{n-1} \mathbb{I}\left[P(s_{t+1}) \ne P(s_t)\right] \text{(Total transitions)}
\end{equation}

\begin{equation}
T^+ = \sum_{t=1}^{n-1} \mathbb{I}\left[P(s_{t+1}) > P(s_t)\right] \text{(Positive transitions)}
\end{equation}

Then:
\begin{equation}
R^+ = \frac{T^+}{T}
\end{equation}

where $\mathbb{I}[\cdot]$ is the indicator function returning 1 if the condition holds, and 0 otherwise.

This metric is computed independently across three dimensions: behavioral states, emotional valence, and cognitive levels. It reflects the proportion of upward movement and thus serves as a compact representation of learning progression and self-regulation effectiveness during instruction.

\begin{table}[ht!]
\centering
\caption{\textbf{Longitudinal positive transition rates of student agent behavior, emotion, and cognition across four instructional sessions.} 
Values are normalized to $[0,1]$; \textbf{Overall} indicates the average across three dimensions.}
\label{tab:longitudinal-transition}
\setlength{\tabcolsep}{6pt}
\small
\begin{tabular}{clcccc}
\toprule
\textbf{Session} & \textbf{Student} & \textbf{B\_Pos} & \textbf{E\_Pos} & \textbf{C\_Pos} & \textbf{Overall} \\
\midrule
\multirow{6}{*}{\textit{1}} 
 & \WangFang & 0.000 & 0.138 & 0.034 & 0.057 \\
 & \ZhangJie & 0.379 & 0.966 & 0.103 & 0.483 \\
 & \ZhangYan & 0.103 & 1.000 & 0.069 & 0.391 \\
 & \LiWei    & 0.759 & 0.966 & 0.172 & 0.632 \\
 & \LiuLi    & 0.000 & 0.034 & 0.000 & 0.011 \\
 & \ZhangTao & 0.034 & 0.207 & 0.000 & 0.080 \\
\midrule
\multirow{6}{*}{\textit{2}} 
 & \WangFang & 0.583 & 0.125 & 0.125 & 0.278 \\
 & \ZhangJie & 0.500 & 1.000 & 0.250 & 0.583 \\
 & \ZhangYan & 0.292 & 1.000 & 0.125 & 0.472 \\
 & \LiWei    & 0.375 & 1.000 & 0.208 & 0.528 \\
 & \LiuLi    & 0.125 & 0.792 & 0.000 & 0.306 \\
 & \ZhangTao & 0.583 & 0.333 & 0.125 & 0.347 \\
\midrule
\multirow{6}{*}{\textit{3}} 
 & \WangFang & 0.500 & 0.567 & 0.067 & 0.378 \\
 & \ZhangJie & 0.533 & 1.000 & 0.167 & 0.567 \\
 & \ZhangYan & 0.867 & 1.000 & 0.367 & 0.744 \\
 & \LiWei    & 0.633 & 1.000 & 0.133 & 0.589 \\
 & \LiuLi    & 0.033 & 0.067 & 0.000 & 0.033 \\
 & \ZhangTao & 0.367 & 0.700 & 0.100 & 0.389 \\
\midrule
\multirow{6}{*}{\textit{4}} 
 & \WangFang & 0.192 & 0.346 & 0.077 & 0.205 \\
 & \ZhangJie & 0.654 & 1.000 & 0.154 & 0.603 \\
 & \ZhangYan & 0.923 & 1.000 & 0.115 & 0.679 \\
 & \LiWei    & 0.192 & 1.000 & 0.115 & 0.436 \\
 & \LiuLi    & 0.269 & 0.231 & 0.038 & 0.179 \\
 & \ZhangTao & 0.385 & 0.308 & 0.077 & 0.256 \\
\bottomrule
\end{tabular}
\end{table}

To complement the main text on long-term evolution, this section applies the above equations to compute the positive transition rate $R^+$ across behavior (B), emotion (E), and cognition (C), as summarized in Tab.~\ref{tab:longitudinal-transition}. Overall, we observe a pattern of \textit{gradual improvement followed by a mild pullback}: many students improve steadily from Session~1 to 3 (e.g., Zhang~Yan: 0.391 $\rightarrow$ 0.744; Wang~Fang: 0.057 $\rightarrow$ 0.378; Zhang~Jie: 0.483 $\rightarrow$ 0.567), whereas Session~4 shows partial regressions under higher cognitive demands (e.g., Wang~Fang: 0.378 $\rightarrow$ 0.205; Zhang~Tao: 0.389 $\rightarrow$ 0.256). We also observe “high start—dip—partial recovery” patterns (e.g., Li~Wei: 0.632 $\rightarrow$ 0.528 $\rightarrow$ 0.589), followed by another decline in Session~4 (0.436).

Dimension-wise, \textbf{behavior} is most sensitive, with several students peaking in Session~2 or 3 (e.g., Zhang~Yan’s B\_Pos: 0.103 $\rightarrow$ 0.867 $\rightarrow$ 0.923). \textbf{Emotion} remains high and stable for many students, reflecting effective teacher regulation (e.g., Zhang~Jie and Li~Wei with E\_Pos = 1.000 in Sessions~2–4), though some show fluctuations (e.g., Wang~Fang: 0.125/0.567/0.346; Liu~Li drops after a high of 0.792). \textbf{Cognition} progresses more slowly with larger variability (e.g., Zhang~Yan’s C\_Pos spikes to 0.367 in Session~3 and softens to 0.115 in Session~4), consistent with the view that cognitive growth requires extended accumulation and reflection.

At the individual level, trajectories reveal personality-consistent stability with interpretable divergence: Zhang~Jie remains high and stable (Overall: 0.483/0.583/0.567/0.603); Zhang~Yan makes a pronounced leap in Session~3 and sustains a high level in Session~4 (0.744/0.679); Li~Wei starts high, dips, then partially recovers (0.632 $\rightarrow$ 0.528 $\rightarrow$ 0.589) before declining again (0.436); Wang~Fang improves then recedes (0.057 $\rightarrow$ 0.378 $\rightarrow$ 0.205); Liu~Li remains low overall with episodic recovery (0.011/0.306/0.033/0.179); and Zhang~Tao shows greater volatility and context sensitivity (0.080/0.347/0.389/0.256). Altogether, $R^+$ provides a compact and interpretable quantification of learning progression and self-regulation, reinforcing the main text’s conclusions on long-term evolution and individual differentiation (see Tab.~\ref{tab:longitudinal-transition}).

\clearpage

\subsection{Multi-Agent Instructional Interaction Cases across Genre-Specific Tasks}

To illustrate how virtual students respond under varying instructional task types, we present representative multi-agent interaction cases from three distinct genres: the lyrical prose \textit{Spring}, the foreign fable \textit{The Emperor’s New Clothes}, and the argumentative essay \textit{Dedication and Joy}. These texts differ in linguistic style, cognitive structure, and affective demand, offering a natural basis for evaluating behavioral variation.

For each task, we sampled three consecutive instructional time steps from the full simulation. We report the complete behavioral logs of selected student agents during these windows, including learning plans, cognitive levels, emotional states, and social behaviors. These interaction snapshots offer micro-level evidence of how agents coordinate cognition, emotion, and participation in real-time.

These examples illustrate how agent behavior adapts to differences in instructional phase, task genre, and teacher–student dynamics. They also provide direct evidence of cross-genre generalizability and personality-driven response stability within the simulation system.

\begin{tcolorbox}[breakable,title={\textbf{Case: Multi-Agent Interaction in "Spring" (Step 8)}}, colback=white, colframe=black!50, breakable]
\textbf{Teacher Content:} “Excellent, everyone. Your understanding and analysis are impressive. Now, I would like to ask: While reading this text, did you notice how Mr. Zhu Ziqing vividly portrayed the beauty of spring through delicate descriptions? For example: ‘Peach trees, apricot trees, and pear trees are all blooming in competition—bright red like fire, pink like clouds, white like snow. The flowers are sweet-scented; close your eyes, and the trees seem already full of peaches, apricots, and pears.’ With these metaphors and personifications, Zhu Ziqing presents spring as a sensory experience. \ZhangYan, could you share how you interpret this passage?”

\textbf{Teacher Emotion:} Encouragement  
\textbf{Teacher Behavior:} Questioning

\textbf{Student Responses:}
\begin{itemize}
  \item \textbf{\WangFang} $\rightarrow$ \textbf{\ZhangYan}: Side talk – “I agree with your point. Zhu Ziqing’s use of personification really helps us feel the vibrancy of spring—it’s like a person waking up, full of hope and energy.” (Emotion: Positive; Cognition: Application)
  \item \textbf{\ZhangJie}: Question – “Teacher, I really love this passage. It feels like he’s painting spring with a brush. The sentence ‘Peach trees, apricot trees, pear trees…’ vividly brings to life the competitive blooming. And ‘the flowers are sweet-scented; close your eyes, and the trees seem already full of fruit’—this conveys not just the beauty, but also the sweetness and hope of spring. I think this is what Zhu Ziqing intended: spring is not just visual, but deeply emotional.” (Emotion: Positive; Cognition: Application)
  \item \textbf{\ZhangYan}: Stand-up response – “I believe Mr. Zhu’s descriptions truly immerse us in the atmosphere of spring. The use of personification and metaphor makes the imagery come alive. That line about blooming trees really shows the vitality of spring.” (Emotion: Positive; Cognition: Application)
  \item \textbf{\LiWei}: Question – “I agree with the teacher and classmates. Zhu’s descriptions are not only vivid but also metaphorical. They reflect the beauty of life and hope. I think this is what he wants us to feel—resilience and positivity even in the face of difficulties. Do others feel the same?” (Emotion: Positive; Cognition: Application)
  \item \textbf{\LiuLi} $\rightarrow$ \textbf{\ZhangTao}: Side talk – “Maybe… I’m not really sure.” (Emotion: Positive; Cognition: Application)
  \item \textbf{\ZhangTao} $\rightarrow$ \textbf{\LiuLi}: Side talk – “Ugh, don’t overthink it. It’s not a big deal anyway.” (Emotion: Negative; Cognition: Understanding)
\end{itemize}

\textbf{Teacher Feedback:} “\LiuLi, great improvement—keep up your active participation. \ZhangJie, your engagement is excellent—looking forward to your insights. \ZhangYan, your communication and understanding are outstanding—keep it up, and remember to listen to others too.”

\end{tcolorbox}

\vspace{0.5em}

\begin{tcolorbox}[breakable,title={\textbf{Case: Multi-Agent Interaction in "Spring" (Step 9)}}, colback=white, colframe=black!50, breakable]
\textbf{Teacher Content:} “Wonderful work, everyone. You’ve gained deep insights into Zhu Ziqing’s text. Now let’s think: How does he express his emotions when portraying spring? Through what language or imagery can we feel his admiration for spring? \WangFang, would you like to share your thoughts first?”

\textbf{Teacher Emotion:} Encouragement  
\textbf{Teacher Behavior:} Questioning

\textbf{Student Responses:}
\begin{itemize}
  \item \textbf{\WangFang}: Stand-up response – “Uh… I think Zhu Ziqing… in ‘Spring’… he uses delicate descriptions to… vividly portray spring. He uses emotional language to describe flowers, trees, rivers… and the word ‘lively’ to reflect his feelings of joy. That’s how I understand his admiration.” (Emotion: Positive; Cognition: Application)
  \item \textbf{\ZhangJie}: Hand raised – “Sure. I think his vivid language makes it feel like we’re witnessing spring with our own eyes. It’s not just about beauty—it’s about his deep love for this season. His writing really impresses me.” (Emotion: Positive; Cognition: Application)
  \item \textbf{\ZhangYan} $\rightarrow$ \textbf{\WangFang}: Side talk – “Exactly, \WangFang. The personification he uses really makes spring feel alive, full of hope.” (Emotion: Positive; Cognition: Application)
  \item \textbf{\LiWei}: Head up, listening, no verbal response (Emotion: Positive; Cognition: Understanding)
  \item \textbf{\LiuLi}: Hand raised – “Uh… I think Zhu Ziqing… really liked spring… his words, um… show he really loved it.” (Emotion: Positive; Cognition: Understanding)
  \item \textbf{\ZhangTao} $\rightarrow$ \textbf{\LiuLi}: Side talk – “\LiuLi, can you explain how Zhu Ziqing shows his love for spring through language and imagery?” (Emotion: Confused; Cognition: Understanding)
\end{itemize}

\textbf{Teacher Feedback:} “\LiuLi, great participation and deep thinking—just be mindful of classroom order. \ZhangYan, well done with your insightful understanding of ‘Spring’—keep up the collaborative spirit.”

\end{tcolorbox}

\vspace{0.5em}

\begin{tcolorbox}[breakable,title={\textbf{Case: Multi-Agent Interaction in "Spring" (Step 10)}}, colback=white, colframe=black!50, breakable]
\textbf{Teacher Content:} “Now let’s examine the structure of ‘Spring’. Zhu Ziqing divides the essay into three parts. First, he begins with anticipation: ‘Looking forward, looking forward, the east wind is coming, and spring’s footsteps are near.’ Then he uses vivid metaphors and delicate imagery to depict spring scenes like blooming trees. Finally, he uses personification—spring is a newborn baby, a little girl, a strong youth—bringing vitality to the season. This layered structure gives the essay rhythm and emotional resonance. \ZhangJie, would you like to share your understanding of this structure?”

\textbf{Teacher Emotion:} Encouragement  
\textbf{Teacher Behavior:} Explanation

\textbf{Student Responses:}
\begin{itemize}
  \item \textbf{\WangFang} $\rightarrow$ \textbf{\ZhangYan}: Side talk – “\ZhangYan… you’re right… Zhu’s personification really shows the vitality and hope of spring!” (Emotion: Positive; Cognition: Application)
  \item \textbf{\ZhangJie}: Stand-up response – “Of course, teacher. I think the structure is brilliant. The beginning sets the tone of anticipation, the middle provides vivid descriptions, and the end gives spring symbolic meaning—energy, rebirth, youth. It’s rhythmic and emotionally deep.” (Emotion: Positive; Cognition: Application)
  \item \textbf{\ZhangYan} $\rightarrow$ \textbf{\WangFang}: Side talk – “I’m glad you felt the vitality too, \WangFang. His words awaken everything—dispelling winter and bringing hope.” (Emotion: Positive; Cognition: Application)
  \item \textbf{\LiWei} $\rightarrow$ \textbf{\ZhangYan}: Side talk – “\ZhangYan, what’s your unique take on Zhu’s metaphors? I love his personification—it inspires our own writing, don’t you think?” (Emotion: Positive; Cognition: Understanding)
  \item \textbf{\LiuLi} $\rightarrow$ \textbf{\ZhangTao}: Side talk – “\ZhangTao, I think Zhu’s vivid words reflect not just spring, but also his love for life and respect for nature.” (Emotion: Positive; Cognition: Understanding)
  \item \textbf{\ZhangTao} $\rightarrow$ \textbf{\LiuLi}: Side talk – “Hmm, I like ‘Back Shadow’ more than ‘Spring’. Have you read that one?” (Emotion: Positive; Cognition: Understanding)
\end{itemize}

\textbf{Teacher Feedback:} “\ZhangJie, wonderful—your structural analysis is spot on. \LiuLi, your input shows a deep grasp of the text—keep it up! \ZhangYan, excellent discussion—maintain your positive attitude and spirit of exploration.”

\end{tcolorbox}

\begin{tcolorbox}[breakable,title={\textbf{Case: Multi-Agent Interaction in "Dedication and Joy" (Step 12)}}, colback=white, colframe=black!50, breakable]
\textbf{Teacher Content:} “Now, let us explore the meaning of 'finding joy in one’s work' (\emph{leye}). It refers to discovering happiness in our work or studies, enjoying what we do. In this way, our tasks become less of a burden and more of a pleasure. How does Liang Qichao interpret \emph{leye}? In his essay, he writes, ‘Be at peace with your dwelling, and find joy in your work’—suggesting that we should experience peace and pleasure in what we do. \ZhangJie, could you share your understanding of \emph{leye}?”

\textbf{Teacher Emotion:} Encouragement  
\textbf{Teacher Behavior:} Questioning

\textbf{Student Responses:}
\begin{itemize}
  \item \textbf{\WangFang}: Note-taking – No verbal response (Emotion: Positive; Cognition: Understanding)
  \item \textbf{\ZhangJie}: Stand-up response – “I’m happy to share my view. I think \emph{leye} means finding joy and value in what we do. As Liang Qichao said, ‘Be at peace with your dwelling, and find joy in your work’—to me, this means experiencing happiness and satisfaction through learning or working. When we love what we do, it becomes our source of joy rather than pressure.” (Emotion: Positive; Cognition: Application)
  \item \textbf{\ZhangYan}: Hand raised – “I think \emph{leye} is about finding enjoyment in our learning or work. As Liang Qichao says, we should find peace and joy in our duties. That way, work becomes something we enjoy. This reflects my experience—I enjoy the learning process and hope others can find that joy too.” (Emotion: Positive; Cognition: Application)
  \item \textbf{\LiWei}: Hand raised – “I believe \emph{leye} is about finding joy and satisfaction in the process of doing things. Liang Qichao’s quote applies not only to work but also to learning. If we find joy in learning, we’ll better understand knowledge and be more motivated. But I wonder, when facing stress or difficulties, how can we still embrace \emph{leye}?” (Emotion: Positive; Cognition: Application)
  \item \textbf{\LiuLi}: Head down sleeping – No response (Emotion: Negative; Cognition: Memory)
  \item \textbf{\ZhangTao} $\rightarrow$ \textbf{\LiuLi}: Chatting – “\LiuLi, did you watch that basketball game yesterday? That dunk was insane!” (Emotion: Negative; Cognition: Memory)
\end{itemize}

\textbf{Teacher Feedback:} “\LiuLi, great to see your improvement—keep participating actively. \ZhangJie, your engagement is excellent. \ZhangYan, you’ve shown strong understanding and communication—keep it up, and remember to listen to your peers.”

\end{tcolorbox}

\vspace{0.5em}

\begin{tcolorbox}[breakable,title={\textbf{Case: Multi-Agent Interaction in "Dedication and Joy" (Step 13)}}, colback=white, colframe=black!50, breakable]
\textbf{Teacher Content:} “Let’s now see how Liang Qichao argues for the importance of \emph{leye}. In his essay ‘Dedication and Joy’, he uses the phrase ‘Be at peace with your dwelling, and find joy in your work’ to explain his perspective. He believes that only when we feel peace and happiness in our duties can we truly understand what \emph{leye} means. The ‘joy’ here refers not only to enjoyment but also to love and appreciation. \ZhangJie raised a great question: is \emph{leye} limited to our work, or can it also apply to learning, life, and hobbies? \LiWei, what do you think?”

\textbf{Teacher Emotion:} Encouragement  
\textbf{Teacher Behavior:} Questioning

\textbf{Student Responses:}
\begin{itemize}
  \item \textbf{\WangFang}: Note-taking – No response (Emotion: Positive; Cognition: Application)
  \item \textbf{\ZhangJie}: Question – “I fully agree with Liang Qichao. He helps us realize that only by loving our work can we find true joy and meaning in it. Then it’s not a burden anymore. But I have a question: can \emph{leye} apply to things beyond work, like learning, life, or hobbies? I think anything we love deeply could be part of \emph{leye}. What do you think, teacher?” (Emotion: Positive; Cognition: Application)
  \item \textbf{\ZhangYan}: Question – “Teacher, I’ve also been thinking about Liang’s view. He’s suggesting that when we genuinely love and enjoy what we do, it becomes a joy. This attitude should apply to learning as well—finding enjoyment rather than just completing tasks. But what if we’re not feeling that joy? How should we adjust our mindset?” (Emotion: Positive; Cognition: Application)
  \item \textbf{\LiWei}: Stand-up response – “Liang’s reasoning is inspiring. He shows that when we love what we do, it stops being pressure—it becomes joy. But I’m a bit confused about ‘being at peace with your dwelling.’ Does that mean we must accept and embrace our current life situation to experience \emph{leye}? I’d love to hear others’ thoughts.” (Emotion: Positive; Cognition: Application)
  \item \textbf{\LiuLi}: Head up listening – No response (Emotion: Positive; Cognition: Understanding)
  \item \textbf{\ZhangTao}: Head up listening – No response (Emotion: Positive; Cognition: Understanding)
\end{itemize}

\textbf{Teacher Feedback:} “\ZhangYan, your thoughtful and active participation is wonderful—keep it up! \ZhangJie, your engagement and depth of understanding help lead the class. \LiWei, your ideas are insightful—continue contributing in discussions!”

\end{tcolorbox}

\vspace{0.5em}

\begin{tcolorbox}[breakable,title={\textbf{Case: Multi-Agent Interaction in "Dedication and Joy" (Step 14)}}, colback=white, colframe=black!50, breakable]
\textbf{Teacher Content:} “Your reflections have been fantastic. \ZhangJie asked a great question—can \emph{leye} apply beyond work, such as to learning or hobbies? I believe \emph{leye} is broad—it applies to any domain where we invest passion. As for \ZhangYan’s question about learning struggles: First, we must accept that difficulties are part of the process. Then, try to see them as challenges and growth opportunities. That’s how we rediscover joy in learning. Does anyone want to share how they’ve done this?”

\textbf{Teacher Emotion:} Encouragement  
\textbf{Teacher Behavior:} Questioning

\textbf{Student Responses:}
\begin{itemize}
  \item \textbf{\WangFang}: Hand raised – “Uh, I… I think when we face learning difficulties… we shouldn’t give up. We… should accept them and… find solutions. If we stay optimistic, we’ll rediscover \emph{leye}.” (Emotion: Positive; Cognition: Application)
  \item \textbf{\ZhangJie}: Stand-up response – “Teacher, I think \emph{leye} is a broad concept—it definitely applies to learning, life, and hobbies. If we commit and love what we do, that’s \emph{leye}. When facing challenges, we should adjust our mindset and find opportunities—that’s how we enjoy learning.” (Emotion: Positive; Cognition: Application)
  \item \textbf{\ZhangYan}: Stand-up response – “I agree. \emph{Leye} means loving not just work but all aspects of life. When we face challenges in learning, changing our perspective helps us turn obstacles into joy. Has anyone had a similar experience? How did you regain your joy in learning?” (Emotion: Positive; Cognition: Application)
  \item \textbf{\LiWei}: Question – “Everyone’s ideas are inspiring. I also believe \emph{leye} is a holistic attitude that includes work, learning, and hobbies. As long as we’re passionate, we’ll find joy and self-worth. Though there may be difficulties, they are growth opportunities. I’d like to hear how others adjusted their mindset during such times.” (Emotion: Positive; Cognition: Understanding)
  \item \textbf{\LiuLi}: Head up listening – No response (Emotion: Positive; Cognition: Understanding)
  \item \textbf{\ZhangTao}: Head up listening – No response (Emotion: Positive; Cognition: Understanding)
\end{itemize}

\textbf{Teacher Feedback:} “\WangFang, great focus and thinking—keep it up, and don’t let anxiety affect your learning. \ZhangYan, your helpfulness and ability to apply learning are admirable—keep going!”

\end{tcolorbox}

\begin{tcolorbox}[title={\textbf{Case: Multi-Agent Interaction in "The Emperor's New Clothes"(Step 12)}}, colback=white, colframe=black!50, breakable]
\textbf{Teacher Content:} “Today, we will be reading \emph{The Emperor’s New Clothes}, a satirical tale by the famous Danish author Hans Christian Andersen. Andersen is one of the most renowned figures in children’s literature, with a diverse range of works spanning fairy tales, novels, poems, and plays. Some of his most beloved stories include \emph{The Little Match Girl}, \emph{The Ugly Duckling}, and \emph{The Little Mermaid}. In his fairy tales, Andersen often reveals human flaws and social injustice from a unique perspective, offering us valuable life lessons.”

\textbf{Teacher Emotion:} Neutral  
\textbf{Teacher Behavior:} Lecture

\textbf{Student Responses:}
\begin{itemize}
  \item \textbf{\WangFang}: Head up listening – No response (Emotion: Positive; Cognition: Understanding)
  \item \textbf{\ZhangJie}: Question – “Teacher, I find this story very interesting. The emperor is tricked by two swindlers who claim that only intelligent people can see their fabric. The emperor never checks the clothes himself but sends others. I wonder—if he had gone himself, would he have seen through the deception? Or if he had the courage to admit he saw nothing, could he have avoided the embarrassment? This story not only mocks vanity and fear of losing face, but also reminds us to have the courage to face the truth. What do others think?” (Emotion: Positive; Cognition: Analysis)
  \item \textbf{\ZhangYan}: Question – “Teacher, this story made me reflect a lot. Andersen’s works always carry deep meaning. Through this tale, he highlights human vanity and blind conformity, showing us the importance of honesty and truth. That’s probably why his works remain so popular—they are both entertaining and meaningful. I also want to ask: does this story still hold relevance today? Don’t we sometimes act like the emperor, pursuing superficial appearances while ignoring real value?” (Emotion: Positive; Cognition: Analysis)
  \item \textbf{\LiWei}: Question – “Teacher, I’ve also thought about what \ZhangJie and \ZhangYan said. To me, this story not only satirizes people who care too much about appearances, but also encourages independent thinking. That’s a key takeaway Andersen wants us to have. I’d love to hear how others interpret it.” (Emotion: Positive; Cognition: Analysis)
  \item \textbf{\LiuLi} $\rightarrow$ \textbf{\ZhangTao}: Chatting – “\ZhangTao, have you watched the latest Marvel movie \emph{Eternals}? I heard the plot is great and the effects are amazing. What did you think?” (Emotion: Negative; Cognition: Memory)
  \item \textbf{\ZhangTao} $\rightarrow$ \textbf{\LiWei}: Chatting – “\LiWei, do you think our school’s football team will win this weekend’s match?” (Emotion: Negative; Cognition: Memory)
\end{itemize}

\textbf{Teacher Feedback:} “\ZhangTao, please stay focused and participate actively so you can better understand the story. \LiuLi, you show strong independent learning ability, but classroom interaction is important too—try joining the discussion!”

\end{tcolorbox}

\vspace{0.5em}

\begin{tcolorbox}[title={\textbf{Case: Multi-Agent Interaction in "The Emperor's New Clothes"(Step 13)}}, colback=white, colframe=black!50, breakable]
\textbf{Teacher Content:} “Students, the questions raised by \ZhangJie, \ZhangYan, and \LiWei were excellent and show deep thinking about the story. Now, can anyone tell me what genre \emph{The Emperor’s New Clothes} belongs to? Is it a novel, a poem, an essay, or a fairy tale?”

\textbf{Teacher Emotion:} Encouragement  
\textbf{Teacher Behavior:} Questioning

\textbf{Student Responses:}
\begin{itemize}
  \item \textbf{\WangFang}: Note-taking – No response (Emotion: Positive; Cognition: Application)
  \item \textbf{\ZhangJie}: Stand-up response – “I think it’s a fairy tale. The plot—like the emperor being deceived—is common in fairy tales. And it teaches us a lesson: to think independently and not blindly follow others.” (Emotion: Positive; Cognition: Analysis)
  \item \textbf{\ZhangYan}: Stand-up response – “I agree—it’s a fairy tale. Although the characters are familiar—like emperors and tailors—the story includes magical elements such as invisible clothes and collective pretending. It also teaches a moral, which is typical of fairy tales.” (Emotion: Positive; Cognition: Application)
  \item \textbf{\LiWei}: Stand-up response – “I also think it’s a fairy tale. It includes fantasy elements and makes us think deeply. But I’d also like to hear what others think.” (Emotion: Positive; Cognition: Application)
  \item \textbf{\LiuLi}: Head down reading – No response (Emotion: Negative; Cognition: Understanding)
  \item \textbf{\ZhangTao} $\rightarrow$ \textbf{\LiuLi}: Chatting – “\LiuLi, what book are you reading? It looks interesting.” (Emotion: Negative; Cognition: Understanding)
\end{itemize}

\textbf{Teacher Feedback:} “\LiuLi, let’s try to stay focused—your intelligence should be used to understand the story. \WangFang, your thoughtful interpretation is excellent—please keep sharing. \ZhangTao, let’s all focus on the lesson—we’re eager to hear your ideas too.”

\end{tcolorbox}

\vspace{0.5em}

\begin{tcolorbox}[title={\textbf{Case: Multi-Agent Interaction in "The Emperor's New Clothes"(Step 14)}}, colback=white, colframe=black!50, breakable]
\textbf{Teacher Content:} “Great work, everyone. \ZhangJie mentioned that if the emperor had the courage to admit he couldn’t see the fabric, he could have avoided embarrassment—a sharp observation. \ZhangYan asked about the story’s modern relevance, and that’s a very valuable question: do we, too, sometimes pursue surface appearances and ignore real value? \LiWei talked about the importance of independent thinking—something we should all cultivate. We’ll continue exploring these ideas in future lessons.”

\textbf{Teacher Emotion:} Encouragement  
\textbf{Teacher Behavior:} Questioning

\textbf{Student Responses:}
\begin{itemize}
  \item \textbf{\WangFang}: Head up listening – No response (Emotion: Positive; Cognition: Analysis)
  \item \textbf{\ZhangJie}: Stand-up response – “I think this story is really meaningful. Like \ZhangJie said, the emperor’s fear of embarrassment kept him from facing the truth. This happens in real life too. The story teaches us to be brave and value independent thinking—just like \LiWei said. I’ll try to apply this understanding in the future and learn from my classmates’ views as well.” (Emotion: Positive; Cognition: Analysis)
  \item \textbf{\ZhangYan}: Stand-up response – “I agree with both \ZhangJie and \LiWei. Courage and critical thinking are crucial. I’d add that in chasing a beautiful life, we shouldn’t lose sight of what’s real. Like the emperor, we risk embarrassment when we chase appearances. This story warns us not to be modern-day emperors.” (Emotion: Positive; Cognition: Analysis)
  \item \textbf{\LiWei}: Stand-up response – “Thank you, teacher. I agree with \ZhangJie’s point about courage and \ZhangYan’s insights on modern relevance. We often do focus too much on appearance and forget real values. This story encourages us to think independently, face truth, and care about substance over form.” (Emotion: Positive; Cognition: Analysis)
  \item \textbf{\LiuLi} $\rightarrow$ \textbf{\ZhangYan}: Chatting – “\ZhangYan, have you seen that popular TV show \emph{Green Fields Under Moonlight}? The characters are so well-written. You should check it out.” (Emotion: Negative; Cognition: Understanding)
  \item \textbf{\ZhangTao} $\rightarrow$ \textbf{\LiWei}: Chatting – “\LiWei, did you notice \LiuLi was chatting during class? Do you know what they were talking about?” (Emotion: Negative; Cognition: Understanding)
\end{itemize}

\textbf{Teacher Feedback:} “\WangFang, great job observing and thinking—don’t be anxious, we’re all learning together. Try sharing your ideas next time. \ZhangYan, your active participation and deep thinking are a real asset to the class. \ZhangJie, your analytical skills and eloquence are excellent—keep up the enthusiasm!”

\end{tcolorbox}

\end{document}

%% file: math_commands.tex
\usepackage{amsmath,amsfonts,bm}

\def\eqref#1{equation~\ref{#1}}

\def\1{\bm{1}}

\DeclareMathAlphabet{\mathsfit}{\encodingdefault}{\sfdefault}{m}{sl}
\SetMathAlphabet{\mathsfit}{bold}{\encodingdefault}{\sfdefault}{bx}{n}

